\documentclass{article}
\usepackage[preprint]{colm2026_conference}

\usepackage{microtype}

\usepackage{amsmath,amsfonts,bm}

\def\eqref#1{equation~\ref{#1}}

\def\1{\bm{1}}

\def\vone{{\bm{1}}}

\DeclareMathAlphabet{\mathsfit}{\encodingdefault}{\sfdefault}{m}{sl}
\SetMathAlphabet{\mathsfit}{bold}{\encodingdefault}{\sfdefault}{bx}{n}

\newcommand{\E}{\mathbb{E}}

\newcommand{\R}{\mathbb{R}}

\newcommand{\softmax}{\mathrm{softmax}}

\newcommand{\KL}{D_{\mathrm{KL}}}

\usepackage[pagebackref=true]{hyperref}
\usepackage{url}

\renewcommand*{\backref}[1]{}
\renewcommand*{\backrefalt}[4]{%
  \ifcase #1 \or {\footnotesize Cited on page~#2.}%
  \else {\footnotesize Cited on pages~#2.}\fi}

\usepackage{amsmath,amssymb,amsthm,mathtools}
\usepackage{bm}
\usepackage{graphicx}
\usepackage{wrapfig}
\usepackage{placeins}
\usepackage{etoc}
\usepackage{booktabs}
\usepackage{capt-of}
\usepackage{array}
\usepackage{enumitem}
\usepackage{xcolor}
\usepackage{colortbl}
\usepackage{tikz}
\usetikzlibrary{positioning,arrows.meta,calc,fit,backgrounds,shapes.geometric,decorations.pathmorphing,decorations.pathreplacing}

\definecolor{plpurple}{HTML}{7C3AED}
\definecolor{plpurplelight}{HTML}{EDE9FE}
\definecolor{plblue}{HTML}{2563EB}
\definecolor{plbluelight}{HTML}{DBEAFE}
\definecolor{plgreen}{HTML}{059669}
\definecolor{plgreenlight}{HTML}{D1FAE5}
\definecolor{plred}{HTML}{DC2626}
\definecolor{plredlight}{HTML}{FEE2E2}
\definecolor{plorange}{HTML}{EA580C}
\definecolor{plorangelight}{HTML}{FFF7ED}
\definecolor{plgray}{HTML}{9CA3AF}
\definecolor{plgraybg}{HTML}{F3F4F6}
\definecolor{plteal}{HTML}{0D9488}
\definecolor{plteallight}{HTML}{CCFBF1}
\definecolor{plframe}{HTML}{D0D0D6}

\colorlet{discretecolor}{plblue}
\colorlet{gaussiancolor}{plorange}
\colorlet{argmaxcolor}{plpurple}
\colorlet{loopcolor}{plteal}
\colorlet{predcolor}{plredlight}
\colorlet{corcolor}{plgreenlight}
\colorlet{grayseq}{plgray}

\hypersetup{colorlinks=true,linkcolor=plpurple,citecolor=plblue,urlcolor=plorange}

\theoremstyle{plain}

\theoremstyle{definition}

\theoremstyle{remark}

\newcommand{\method}{\textsc{Duo}}
\newcommand{\methodpp}{\texorpdfstring{\mbox{\textsc{Duo}\textsuperscript{++}}}{Duo++}}

\newcommand{\onehotset}{\mathcal{V}}

\newcommand{\x}{\mathbf{x}}
\newcommand{\z}{\mathbf{z}}
\newcommand{\w}{\mathbf{w}}

\newcommand{\mask}{\bm{m}}
\newcommand{\beps}{\bm{\epsilon}}

\newcommand{\Tmap}{{\color{argmaxcolor}\mathcal{T}}}
\newcommand{\at}{\alpha_t}
\newcommand{\as}{\alpha_s}
\newcommand{\atg}{\tilde{\alpha}_t}

\newcommand{\cat}{\mathrm{Cat}}
\newcommand{\denoise}{\mathbf{x}_{\theta}}
\newcommand{\nelbo}{\mathcal{L}_{\text{NELBO}}}
\newcommand{\loops}{S}
\newcommand{\loopmax}{S_{\max}}

\newcommand{\midblock}{\mathcal{B}}

\newcommand{\nexec}{n_{\mathrm{exec}}}
\newcommand{\Ffwd}{F_{\mathrm{fwd}}}
\newcommand{\Ftrain}{F_{\mathrm{train}}}
\newcommand{\Fbase}{F_{\mathrm{base}}}
\newcommand{\Floop}{F_{\mathrm{loop}}}
\newcommand{\Flayer}{F_{\mathrm{layer}}}
\newcommand{\Es}{\mathbb{E}[\loops]}
\newcommand{\xL}{\x^{1:L}}
\newcommand{\ztL}{\z_t^{1:L}}

\newcounter{finding}
\newcommand{\findingbox}[1]{%
  \refstepcounter{finding}%
  \par\smallskip
  \noindent\begin{tikzpicture}
    \node[
      draw=plpurple,
      fill=plpurplelight!35,
      rounded corners=5pt,
      line width=0.8pt,
      inner xsep=8pt,
      inner ysep=6pt,
      text width=\dimexpr\linewidth-18pt\relax,
      align=left
    ]{\small\textbf{Finding \thefinding.} #1};
  \end{tikzpicture}%
  \par\smallskip
}

\definecolor{darkblue}{rgb}{0, 0, 0.5}
\hypersetup{colorlinks=true, citecolor=darkblue, linkcolor=darkblue, urlcolor=darkblue}

\title{Looped Uniform Diffusion Language Models}

\author{Saiteja Utpala\textsuperscript{1} \And
Chieh-Hsin Lai\textsuperscript{2} \And
Yuki Mitsufuji\textsuperscript{2} \And
Praneeth Vepakomma\textsuperscript{1,3} \AND
\normalfont\textsuperscript{1}MBZUAI\qquad
\textsuperscript{2}Sony Group Corporation\qquad
\textsuperscript{3}MIT}

\newcommand{\codeurl}{https://github.com/SaitejaUtpala/looped_udlm}
\newcommand{\checkpointsurl}{https://huggingface.co/saitejautpala/looped_udlm}
\newcommand{\logsurl}{https://wandb.ai/saitejautpala/looped_udlm}

\definecolor{linkslate}{HTML}{334155}
\newcommand{\paperlink}[3]{%
  \href{#2}{\raisebox{-0.16em}{\includegraphics[height=0.95em]{#1}}%
  \hspace{0.45em}#3}}

\begin{document}

\maketitle
\lhead{Preprint}

\vspace*{-1.8em}
\begin{center}
  \hypersetup{urlcolor=linkslate}\color{linkslate}\small\sffamily
  \paperlink{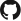}{\codeurl}{Code}\hspace{2.4em}%
  \paperlink{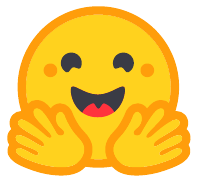}{\checkpointsurl}{Checkpoints}\hspace{2.4em}%
  \paperlink{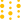}{\logsurl}{Training logs}
\end{center}
\vspace{0.9em}

\begin{abstract}
Uniform diffusion language models (UDLMs) generate text by iteratively denoising sequences of ordinary vocabulary tokens, any of which may be corrupted and subsequently revised. During training, UDLMs must denoise from a context that may itself be unreliable, whereas masked diffusion language models (MDLMs) always condition on a clean context. This challenging training objective makes UDLMs more FLOP-intensive to train than MDLMs, but it yields stronger downstream reasoning performance. Recently, selective looping, which recursively applies a few early Transformer layers within each denoising step, has been shown to improve training efficiency in MDLMs under an iso-FLOP budget. We investigate looping in UDLMs while keeping the training objective and sampler unchanged. We find that looping significantly improves training efficiency: looped UDLMs reach the validation NLL of a non-looped UDLM trained on about 0.5 trillion tokens with up to \(2.8\times\) fewer training FLOPs on OpenWebText and \(3.3\times\) fewer on FineWeb-Edu. We also find that, at a matched NFE budget, looped UDLMs reach up to \(9.8\%\) lower generation perplexity than the non-looped UDLM, with most gains achieved at moderate loop counts. Finally, we study different loop-count distributions and find that sampling smaller loop counts more frequently during training improves inference efficiency: on downstream tasks, this strategy uses \(12\%\) fewer loops than uniform loop-count sampling without degrading accuracy.
\end{abstract}

\setcounter{topnumber}{1}
\etocdepthtag.toc{main}
\section{Introduction}
\label{sec:intro}
\begin{figure}[t]
  \centering
  \includegraphics[width=\textwidth]
    {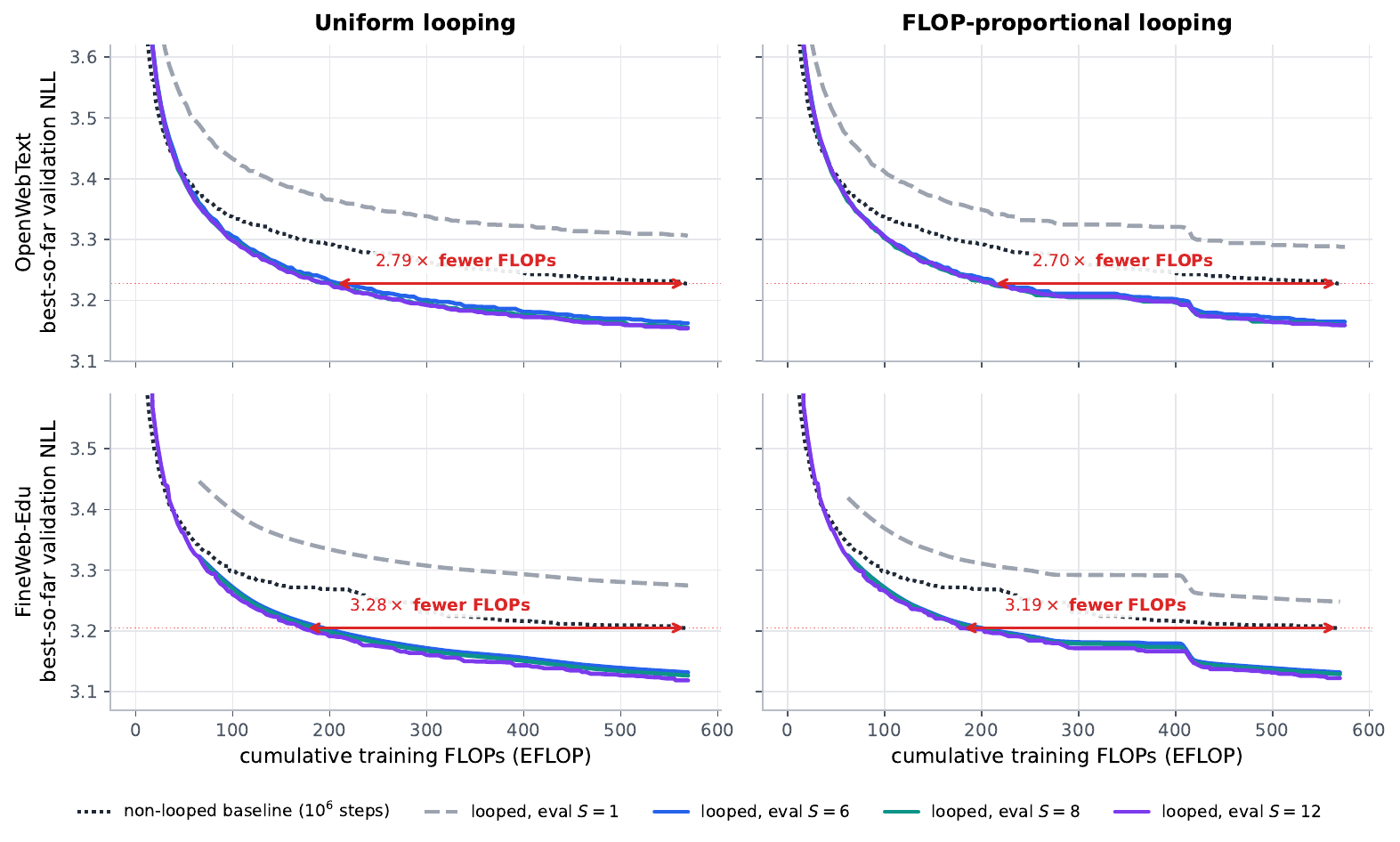}
  \caption{\textbf{Looping saves training compute on OWT and FineWeb-Edu.} Best validation NLL so far against training FLOPs for uniform (left) and FLOP-proportional (right) looping, on OWT (top) and FineWeb-Edu (bottom).}
  \label{fig:training-elbow}
\end{figure}

Discrete diffusion language models (dLLMs) generate text by starting from a
corrupted sequence and denoising many positions in
parallel~\citep{austin2021structured,lou2024discrete,sahoo2024simple,nie2025large,lai2025principles}.
The forward process decides whether the denoiser knows which tokens are
corrupted. \emph{Masked} diffusion language
models (MDLMs) replace corrupted tokens with a dedicated mask symbol, so the
corrupted positions are always known. Under the standard absorbing-state
reverse process, unmasked tokens are never
changed~\citep{sahoo2024simple,shi2025simplified,ou2024RADD,zheng2025masked}.
\emph{Uniform} diffusion language models (UDLMs) instead replace tokens with
random tokens from the vocabulary, so every position holds an ordinary token
that gives no sign of being
corrupted~\citep{austin2021structured,schiff2025simple,sahoo2025diffusionduality,hong2026does}.
Because of this, the reverse process can revise any position at any denoising
step. The distinction is becoming more important as uniform diffusion is
scaled to billion-parameter language models for
reasoning~\citep{diffusiongemmateam2026diffusiongemmatechnicalreport,sahoo2026unlockinglosslessspeedupsllms}.

Denoising in UDLMs is harder than in MDLMs because the denoiser must predict corrupted tokens from a context that may itself be corrupted. If a token does not fit its context, either the token or one of its neighbors was corrupted, and the denoiser has to work out which. At each position it therefore predicts two things. The first is the probability that the observed token was retained, which depends on the noise level and on how well the token agrees with its context. The second is the clean token itself, in case it was
replaced. Both are read from the model's contextual prediction of the clean token, so refining that prediction should improve both. The most
straightforward way to refine it is to scale up the model by adding layers, but this increases the parameter count. Another way is \emph{looping}, which spends more computation on the same
noisy sequence by reusing existing layers, without adding
parameters~\citep{dehghani2018universal,lan2019albert,giannou2023looped,geiping2025scaling}.
\citet{lee2026looped} recently showed that selectively looping a small block
of early-middle layers improves training efficiency in MDLMs, where
unresolved positions are explicitly identified. Whether looping also helps
UDLMs, where the denoiser cannot tell which positions are corrupted, remains
open.
We therefore ask:
\emph{How useful is looping for UDLM denoising, and how does the
loop-count distribution affect training and inference?}

To investigate this question, we integrate looping into UDLMs while keeping the training objective and sampler unchanged, using
\methodpp{}~\citep{deschenaux2026diffusion}, the most efficient recipe for training UDLMs in the literature. Besides the uniform loop-count distribution used in prior work on looped masked diffusion~\citep{lee2026looped}, we study two alternatives: FLOP-proportional sampling, which gives each loop count the same expected share of training FLOPs, so smaller loop counts are sampled more often, and noise-proportional sampling, which assigns larger loop counts to noisier inputs. At inference, we consider fixed looping, which uses the same loop count at every denoising step, and adaptive looping, which halts once the relative change in the hidden state between consecutive iterations falls below a threshold.
\paragraph{Main findings.}
We summarize our observations as follows:
\begin{enumerate}[leftmargin=1.4em,itemsep=2pt,topsep=2pt]
\item \textbf{Looping makes UDLM training more compute-efficient.}
Looped UDLMs reach the validation NLL of a non-looped UDLM trained on
about \(0.5\) trillion tokens with up to \(2.8\times\) fewer training FLOPs on OpenWebText~\citep{Gokaslan2019OpenWeb} and \(3.3\times\) fewer on FineWeb-Edu~\citep{penedo2024fineweb}.

\item \textbf{Looping helps at inference.} Looped UDLMs reach up to \(9.8\%\) lower generative perplexity than the non-looped UDLM at the same number of denoising steps, with most of the gain captured at moderate loop counts. On seven out-of-domain datasets, looping at inference lowers the zero-shot perplexity of the non-looped UDLM by \(11.5\%\) on average.

\item \textbf{Loops mainly refine predictions at replaced tokens.} Extra loop iterations mostly improve the predicted clean token at replaced positions, not the choice of which tokens to keep. The gain is largest at intermediate noise.

\item \textbf{Adaptive looping saves loop iterations without losing accuracy.} On ten zero-shot multiple-choice tasks, adaptive looping matches the accuracy of twelve fixed loop iterations while running \(26.0\%\) fewer of them with uniform sampling and \(34.8\%\) fewer with FLOP-proportional sampling.
\end{enumerate}
\section{Background}
\label{sec:background}

\begin{figure}[t]
\centering
\definecolor{figborder}{HTML}{D8DDE6}
\definecolor{figheadbg}{HTML}{F5F7FB}
\definecolor{figtitle}{HTML}{334155}
\definecolor{figink}{HTML}{111827}
\definecolor{figmut}{HTML}{6B7280}
\definecolor{figacc}{HTML}{4F46E5}
\definecolor{figaccbg}{HTML}{EEF1FE}
\definecolor{figaccbd}{HTML}{A9AEEA}
\resizebox{0.95\textwidth}{!}{%
\begin{tikzpicture}[
  font=\small,
  code/.style={anchor=west, font=\small\bfseries\boldmath, text=figink},
  num/.style={anchor=west, font=\scriptsize\ttfamily, text=figmut},
  glab/.style={anchor=east, font=\tiny\sffamily, text=figmut},
  title/.style={anchor=west, font=\small\sffamily\fontseries{bx}\selectfont, text=figtitle},
  tag/.style={anchor=east, font=\scriptsize\sffamily\fontseries{bx}\selectfont, text=figmut},
  chip/.style={rounded corners=2.5pt, fill=figaccbg, draw=figaccbd,
    line width=0.8pt, text=figacc, inner xsep=5pt, inner ysep=2.6pt,
    font=\small\bfseries\boldmath},
  mtag/.style={rounded corners=2.5pt, draw=figborder, line width=0.8pt,
    fill=white, text=figtitle, inner xsep=5pt, inner ysep=2pt,
    font=\scriptsize\sffamily\fontseries{bx}\selectfont},
]
\def\hy{7.55} 
\begin{scope}
\begin{scope}
  \clip[rounded corners=7pt] (0,2.95) rectangle (7.5,8.3);
  \fill[white] (0,2.95) rectangle (7.5,8.3);
  \fill[figheadbg] (0,\hy) rectangle (7.5,8.3);
\end{scope}
\draw[figborder, line width=0.9pt, rounded corners=7pt] (0,2.95) rectangle (7.5,8.3);
\draw[figborder, line width=0.9pt] (0,\hy) -- (7.5,\hy);
\node[title] at (0.3,7.92) {FORWARD PROCESS \; + \; TRAIN};
\node[code] at (0.3,7.20) {\(\xL\leftarrow\mathrm{batch}(q_{\rm data})\)};
\node[code] at (0.3,6.70) {\(t\sim\mathcal U[0,1],\ \ \beps\sim\mathcal N(\mathbf0,\bm I)\)};
\node[code] at (0.3,6.20) {\(\w_t\leftarrow\atg\,\xL+\sqrt{1-\atg^2}\;\beps\)};
\node[code] at (0.3,5.70) {\(\at\leftarrow\Tmap(\atg)\)};
\node[code] at (0.3,5.20) {\(\ztL\leftarrow\mathbf e_{\arg\max_j w_{t,j}}\)};
\node[code] at (0.3,4.70) {\(\mathbf y_t\leftarrow\operatorname{TopKSoftmax}_{\tau,k}(\w_t)\)};
\node[chip, anchor=west] (traincall) at (0.3,4.18)
  {\(\widehat{\x}\leftarrow\textsc{LoopedDenoise}(\mathbf y_t,t,\loops)\)};
\node[code] at (0.3,3.62)
  {\(\mathcal L_{\rm tok}\leftarrow\textstyle L^{-1}\sum_\ell f_{\rm Duo}(\z_t^\ell,\widehat{\x}^{\,\ell},\at;\x^\ell)\)};
\end{scope}

\begin{scope}
\begin{scope}
  \clip[rounded corners=7pt] (8.0,2.95) rectangle (15.5,8.3);
  \fill[white] (8.0,2.95) rectangle (15.5,8.3);
  \fill[figheadbg] (8.0,\hy) rectangle (15.5,8.3);
\end{scope}
\draw[figborder, line width=0.9pt, rounded corners=7pt] (8.0,2.95) rectangle (15.5,8.3);
\draw[figborder, line width=0.9pt] (8.0,\hy) -- (15.5,\hy);
\node[title] at (8.3,7.92) {REVERSE PROCESS};
\node[code] at (8.3,7.20) {\(\z_1^\ell\stackrel{\rm iid}{\sim}\cat(\vone/K)\)};
\node[code] at (8.3,6.78) {\(\mathbf{for}\ n=N,\ldots,1:\)};
\node[code] at (8.75,6.38) {\(t\leftarrow t_n,\ \ s\leftarrow t_{n-1}\)};
\node[chip, anchor=west] (samplecall) at (8.75,5.88)
  {\(\widehat{\x}\leftarrow\textsc{LoopedDenoise}(\ztL,t,\loops(t))\)};
\node[code] at (8.75,5.36)
  {\(r_\theta^\ell\leftarrow q_{0\mid t}(\cdot\mid\z_t^\ell,\x^\ell\!\leftarrow\!\widehat{\x}^{\,\ell})\)};
\node[code] at (8.75,4.84)
  {\(p_{\rm pred}^\ell\leftarrow q_{s\mid t}(\cdot\mid\z_t^\ell,\x^\ell\!\leftarrow\!\widehat{\x}^{\,\ell})\)};
\node[code] at (8.75,4.32)
  {\(p_{\rm cor}^\ell\leftarrow\alpha_s r_\theta^\ell+(1-\alpha_s)\mathbf u\)};
\node[code] at (8.75,3.80)
  {\(p_\Psi^\ell\leftarrow\kappa_t p_{\rm pred}^\ell+(1-\kappa_t)p_{\rm cor}^\ell\)};
\node[code] at (8.75,3.28)
  {\(\z_s^\ell\sim\cat(p_\Psi^\ell)\)};
\end{scope}

\begin{scope}[yshift=0.05cm]
\begin{scope}
  \clip[rounded corners=7pt] (0,0.5) rectangle (15.5,2.55);
  \fill[white] (0,0.5) rectangle (15.5,2.55);
  \fill[figaccbg] (0,1.95) rectangle (15.5,2.55);
\end{scope}
\draw[figaccbd, line width=0.9pt, rounded corners=7pt] (0,0.5) rectangle (15.5,2.55);
\draw[figaccbd, line width=0.9pt] (0,1.95) -- (15.5,1.95);
\node[anchor=west, font=\small\sffamily\fontseries{bx}\selectfont, text=figacc] at (0.3,2.25)
  {{\rmfamily\textsc{LoopedDenoise}}\((\mathbf y_t,t,\loops)\)};
\node[code] at (0.3,1.58)
  {\(\mathbf e_t\leftarrow E_\theta(\mathbf y_t),\quad
     \mathbf h^{(0)}\leftarrow H_\theta(\mathbf e_t,t)\)};
\node[code] at (0.3,1.13)
  {\(\mathbf{for}\ r=0,\ldots,\loops-1:\ \ \mathbf h^{(r+1)}\leftarrow M_\theta(\mathbf h^{(r)},t)\)};
\node[code] at (0.3,0.68)
  {\(\mathbf{return}\ \widehat{\x}=\softmax\!\big(T_\theta(\mathbf h^{(\loops)})\big)\)};
\end{scope}
\end{tikzpicture}%
}%
\caption{\textbf{Looped UDLM pipeline.} Left: forward process and training. Right: reverse process with the $\Psi$ sampler. $L$ is the sequence length, $K$ the vocabulary size, and $\loops$ the number of loop iterations.}
\label{fig:duo}
\end{figure}
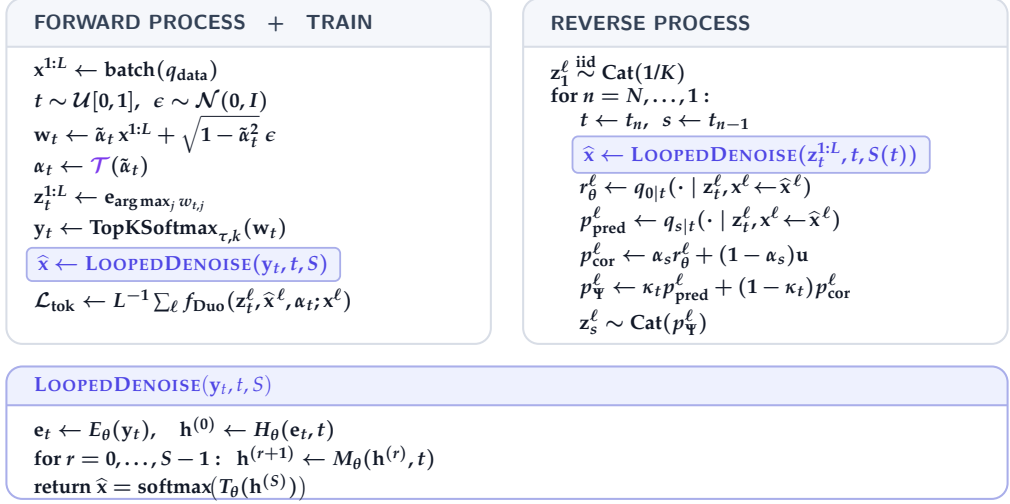

We briefly review uniform diffusion language models (UDLMs) and the
\method{}/\methodpp{} training and sampling framework used in our
experiments.
Further technical details are provided in Appendix~\ref{app:primer}.

\subsection{Uniform diffusion language models}
\label{sec:bg-usdm}

A UDLM corrupts a sequence by replacing some of its tokens with random
tokens from the vocabulary~\citep{austin2021structured,schiff2025simple}. Let \(\xL=(\x^1,\ldots,\x^L)\sim q_{\mathrm{data}}\) be a
clean sequence of \(L\) tokens from a vocabulary of size \(K\),
represented as one-hot vectors in \(\R^K\). At time
\(t\in[0,1]\), each token is retained with probability \(\at\) and otherwise
replaced by a uniformly drawn token:
\begin{equation}
\label{eqn:marginal}
q_t(\ztL\mid\xL)
=\prod_{\ell=1}^{L}
\cat\!\left(\z_t^\ell;\at\x^\ell+(1-\at)\mathbf u\right),
\qquad \mathbf u=\vone/K,\quad \alpha_0=1,\quad\alpha_1\simeq0.
\end{equation}
Here \(\cat(\cdot;\mathbf p)\) denotes a categorical distribution with
probabilities \(\mathbf p\). So \(t=0\) gives clean data and \(t=1\) gives
nearly uniform noise. A corrupted position can even draw its original token again, and no marker shows which positions were corrupted. Equation~\ref{eqn:marginal} corrupts positions independently, while the clean tokens can depend on each other.

To generate, a denoiser uses the entire noisy sequence to predict a clean-token
distribution at every position,
\(\widehat{\x}=\denoise(\ztL,t)\in(\Delta^K)^L\), where \(\Delta^K\)
is the \(K\)-category probability simplex. These contextual predictions
are used in the known Bayesian reverse transitions to a less noisy time
\(s<t\). Each transition updates all positions in parallel.
\method{}~\citep{sahoo2025diffusionduality} improves UDLM training
convergence, and \methodpp{}~\citep{deschenaux2026diffusion} further
improves on \method{} by reducing peak memory, as we review next.

\subsection{\method{} and \methodpp{}}
\label{sec:bg-duo}

A categorical token records only the winning vocabulary entry. To provide
richer training inputs, \method{}~\citep{sahoo2025diffusionduality} obtains each noisy token from an auxiliary
Gaussian vector. For a single token, suppressing the position index,
\begin{equation}
\label{eqn:duality}
\w_t=\atg\x+\sqrt{1-\atg^2}\,\beps,\quad
\z_t=G(\w_t):=\mathbf e_{\arg\max_j w_{t,j}},\quad
\at=\Tmap(\atg),\quad
\beps\sim\mathcal N(\mathbf0,\bm I).
\end{equation}
With \(\at=\Tmap(\atg)\), the winning token \(G(\w_t)\) follows exactly
the UDLM marginal in Equation~\ref{eqn:marginal}. The correspondence holds
only for the marginals at each time. The Gaussian and categorical
diffusion trajectories remain different. While the hard token
\(G(\w_t)\) defines the corruption target, the denoiser receives the
positionwise soft inputs \(\mathbf y_t^\ell=\softmax(\w_t^\ell/\tau)\),
where \(\tau>0\) is a temperature. These retain the margins between competing
tokens that \(\arg\max\) discards, and the hard-token limit is recovered as
\(\tau\downarrow0\). \method{} uses these soft inputs as a curriculum for the
first half of training and hard tokens for the second half. This lowers gradient variance, and the model converges about twice
as fast~\citep{sahoo2025diffusionduality}.
\methodpp{}~\citep{deschenaux2026diffusion} approximates the soft inputs with
only the largest \(k\) Gaussian entries and an estimate of the remaining
softmax mass, so that the curriculum can be used with large
vocabularies. In practice \(k=2\) already works
well, reducing peak memory by \(33\%\) and training time by \(25\%\)
relative to \method{}~\citep{deschenaux2026diffusion}.

Both methods use the inherited UDLM rate-based loss, always evaluated at the
hard corruption target \(\z_t^\ell=G(\w_t^\ell)\). During the curriculum, the
denoiser computes its contextual predictions from the soft inputs. Afterwards,
it receives the hard tokens:
\begin{equation}
\label{eqn:nelbo}
\mathcal L_{\mathrm{tok}}
=\E_{\xL,t,\w_t^{1:L}}\!\left[
  \frac1L\sum_{\ell=1}^{L} f_{\mathrm{Duo}}^\ell\right].
\end{equation}
Here \(f_{\mathrm{Duo}}^\ell\) compares the true and predicted reverse-transition
rates at position \(\ell\). With hard denoiser inputs, this is the token-normalized
\emph{negative evidence lower bound} (NELBO), an upper bound on negative
log-likelihood. Soft inputs give a biased curriculum surrogate. The exact rate
and derivation appear in Appendix~\ref{app:primer}.

\paragraph{Categorical sampling.}
Generation uses hard categorical tokens rather than Gaussian inputs.
The denoiser's clean-token predictions are substituted into the UDLM Bayesian
posterior to obtain an ancestral predictor \(p_{\mathrm{pred}}^\ell\).
A corrector \(p_{\mathrm{cor}}^\ell\) re-noises the inferred clean-token
distribution instead.
The \(\Psi\)-sampler~\citep{deschenaux2026diffusion} mixes them:
\begin{equation}
\label{eqn:bg-psi}
p_\Psi^\ell
=\kappa_t p_{\mathrm{pred}}^\ell+(1-\kappa_t)p_{\mathrm{cor}}^\ell,
\qquad \z_s^\ell\sim\cat(\cdot;p_\Psi^\ell),\quad\kappa_t\in[0,1].
\end{equation}
The update is applied in parallel across positions and repeated toward
clean time. Figure~\ref{fig:duo} summarizes both pipelines. Our work changes
only the denoiser computation, leaving the loss and sampler unchanged.

\section{Looped Uniform Diffusion Models}
\label{sec:method}

The corruption process of UDLMs makes training challenging, because tokens
must be predicted from an unreliable context.
Concretely, a UDLM needs more training FLOPs than an MDLM to reach the same
validation NLL~\citep{sahoo2026scaling}. Recent works have shown that looping,
which repeatedly applies a shared set of Transformer layers, adds depth
without adding parameters~\citep{dehghani2018universal,giannou2023looped,geiping2025scaling}
and improves the training efficiency of masked diffusion language models at
matched training FLOPs~\citep{lee2026looped}. Motivated by this, we
study the effect of looping on the training efficiency and downstream
performance of UDLMs, keeping the corruption process, the \methodpp{}
objective, and the sampler unchanged.
\S\ref{sec:method-reliability} quantifies the uncertainty created by
uniform corruption, and \S\ref{sec:method-loop} introduces the looped UDLM
denoiser and the loop-count distributions used to train it.

\subsection{Unmarked corruption and latent reliability}
\label{sec:method-reliability}

\paragraph{An exact reliability decomposition.}
Both MDLMs and UDLMs can be described by a \emph{keep-or-replace}
corruption kernel:
\begin{equation}
\label{eqn:general-corruption}
q_t(z\mid x)
=\at\mathbf 1\{z=x\}+(1-\at)\nu(z),
\end{equation}
where \(\nu\) is a replacement distribution independent of the
clean token, possibly supported on an augmented vocabulary.
In absorbing MDLMs \(\nu=\delta_{\mask{}}\), and in UDLMs \(\nu\) is uniform
over the ordinary vocabulary. Let \(C_t^\ell\) be the latent reliability indicator, which is \(1\)
for a retained token and \(0\) otherwise. For an observed
token \(\z_t^\ell=\mathbf e_i\), let
\(\pi_t^\ell(j):=\Pr(\x^\ell=\mathbf e_j\mid\z_t^{-\ell},t)\) denote the
distribution of the clean token given the rest of the noisy sequence,
referred to as the contextual prediction. A retained token equals its clean
value and a replaced one carries no information about it, so Bayes'
rule gives
\begin{equation}
\label{eqn:reliability-decomposition}
\begin{aligned}
\gamma_t^\ell
:=\Pr(C_t^\ell=1\mid\ztL,t)
&=\frac{\at\,\pi_t^\ell(i)}
       {\at\,\pi_t^\ell(i)+(1-\at)\nu(\mathbf e_i)},\\
\Pr(\x^\ell=\mathbf e_j\mid\ztL,t)
&=\gamma_t^\ell\,\mathbf 1\{j=i\}
 +(1-\gamma_t^\ell)\pi_t^\ell(j),
\end{aligned}
\end{equation}
for \(0<\at<1\). The clean-token posterior is thus a mixture of a point
mass on the observed token and the contextual prediction, weighted by
the retention probability \(\gamma_t^\ell\). For UDLMs,
\(\nu(\mathbf e_i)=1/K\) and
\begin{equation}
\label{eqn:udm-reliability}
\frac{\gamma_t^\ell}{1-\gamma_t^\ell}
=\frac{K\at}{1-\at}\,\pi_t^\ell(i),
\end{equation}
so retention is not marked in the input: even at a fixed diffusion
time, it has to be inferred from how well the token fits its context.
For MDLMs, the same weight is \(1\) for an ordinary token and \(0\) for
a mask. Because \(\pi_t^\ell\) is computed from positions that may
themselves have been replaced, these judgments are coupled across
positions~\citep{hong2026does}. Appendix~\ref{app:reliability} gives
the derivations, and related identities appear in
\citet{gourevitch2026uniform}.

\paragraph{What the \methodpp{} denoiser estimates.}
The UDLM NELBO (Equation~\ref{eqn:nelbo}) depends on the denoiser output
\(\widehat\x^\ell\) only through \(K\at\widehat\x^\ell+(1-\at)\vone\),
and with hard inputs it is minimized at \(\widehat\x^\ell=\pi_t^\ell\)
(Appendix~\ref{app:reliability}). A trained \methodpp{} denoiser
therefore estimates the contextual prediction, and
Equation~\ref{eqn:udm-reliability} turns its output into an estimated
retention probability \(\hat\gamma_t^\ell\). Since additional loop iterations change only
\(\widehat\x^\ell\), we use the decomposition in
Equation~\ref{eqn:reliability-decomposition} to understand how looping
helps in \S\ref{sec:zero-shot-ppl}.

\subsection{Looped denoiser and loop-count training}
\label{sec:method-loop}

We loop a slice of middle blocks of the \methodpp{} denoiser and draw a
random loop count at every training step. Besides uniform sampling, we
introduce two new loop-count distributions: one that balances training
FLOPs across loop counts, and one that ties the loop count to the noise
level.

\begin{figure}[t]
  \centering
  \begin{minipage}[b]{0.47\textwidth}\centering
    \includegraphics[width=\linewidth]
      {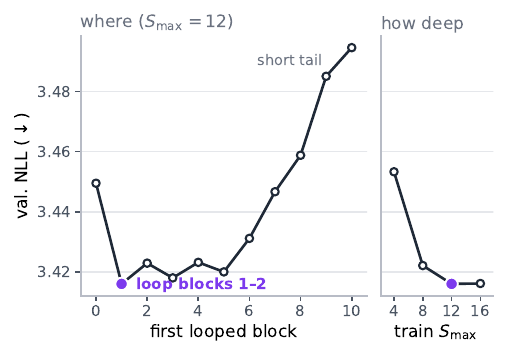}\\
    {\small (a) where to loop and how deep to train}
  \end{minipage}\hfill
  \begin{minipage}[b]{0.47\textwidth}\centering
    \includegraphics[width=\linewidth]
      {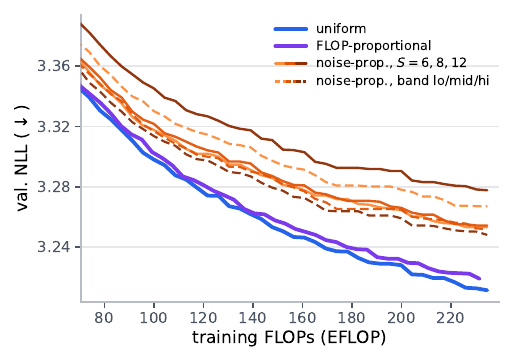}\\
    {\small (b) noise-proportional distribution}
  \end{minipage}
  \caption{\textbf{Design choices behind the looped models.} (a)~NLL at $\loops=8$ for each starting block of the looped slice (left) and best NLL for each training cap $\loopmax$ (right). Purple marks the setting used in the paper. (b)~All six noise-proportional evaluations are worse than uniform and FLOP-proportional at the same FLOPs.}
  \label{fig:loop-layer-ablation}
  \label{fig:noiseprop-zoom}
\end{figure}

\paragraph{Architecture.}
The \methodpp{} denoiser is a Diffusion Transformer
(DiT)~\citep{peebles2023scalable} with \(n\) time-conditioned blocks,
which we split into a head \(H_\theta\) of \(n_H\) blocks, a slice of
middle blocks \(M_\theta\) with \(n_\midblock\) blocks, and a tail
\(T_\theta\) of \(n_T\) blocks that also holds the output projection.
The input \(\mathbf y_t\) follows the conventions of
\S\ref{sec:bg-duo}: soft or top-\(k\) inputs during the curriculum and
hard tokens otherwise. It is embedded once by \(E_\theta\), passed
through the head, looped through the middle slice for \(\loops\)
iterations, and mapped to clean-token predictions by the tail:
\begin{equation}
\label{eqn:loop}
\begin{aligned}
\mathbf e_t&=E_\theta(\mathbf y_t),&
\mathbf h^{(0)}&=H_\theta(\mathbf e_t,t),\\
\mathbf h^{(r+1)}
&=M_\theta(\mathbf h^{(r)},t),
&&r=0,\ldots,\loops-1,\\
\widehat{\x}
&=\softmax\!\bigl(T_\theta(\mathbf h^{(\loops)})\bigr).
\end{aligned}
\end{equation}
The slice $M_\theta$ uses the same weights at every pass, so looping adds
depth but no parameters. With $\loops$ passes the network has
$n_H+\loops n_\midblock+n_T$ blocks, and $\loops=1$ is the original network.
The noisy input stays fixed during the loop and changes only between
diffusion steps (Figure~\ref{fig:duo}).

\paragraph{Stochastic loop-count training.}
At each optimizer step, we sample a loop count \(\loops\) from a
distribution \(p(\loops)\) over \(\{1,\ldots,\loopmax\}\), unroll the
slice \(\loops\) times, and backpropagate the \methodpp{} objective
(Equation~\ref{eqn:nelbo}) through every iteration. This gives one model
that can be run with different loop counts at inference, either fixed or
chosen adaptively at each diffusion step (\S\ref{sec:setup}). The choice of \(p(\loops)\)
matters for the training budget, because the modeled per-sequence
training cost grows with the loop count:
\begin{equation}
\label{eqn:step-cost}
F(\loops)=\Fbase\bigl(1+(\loops-1)\rho\bigr),
\qquad
\rho=\frac{n_\midblock\Flayer}{\Fbase},
\end{equation}
where \(\Fbase\) is the non-looped cost and \(\Flayer\) is the cost of
one block in the looped slice (Appendix~\ref{app:flops}).

\paragraph{(a) Uniform.}
LoopMDM~\citep{lee2026looped} samples
\(\loops\sim\mathcal U\{1,\ldots,\loopmax\}\), and we
use this as our default with \(\loopmax=12\). Each loop count is then trained equally
often. Since a step with more loops costs more
(Equation~\ref{eqn:step-cost}), most of the training FLOPs go to large loop
counts, even though the inference gains level off at moderate loop counts
(\S\ref{sec:fixed-loop-results}).

\paragraph{(b) FLOP-proportional.}
To balance training FLOPs instead of update counts, we sample each loop
count with probability inversely proportional to its cost,
\begin{equation}
\label{eqn:flopprop}
p_{\rm fp}(\loops)
=\frac{F(\loops)^{-1}}
       {\sum_{r=1}^{\loopmax}F(r)^{-1}},
\qquad
p_{\rm fp}(\loops)F(\loops)=\mathrm{const},
\end{equation}
which adds no tunable parameter beyond the architecture-defined costs
and \(\loopmax\), and reduces to uniform sampling when \(\rho=0\). Under
\(p_{\rm fp}\), the expected cost of a training step is the harmonic mean
of \(F(1),\ldots,F(\loopmax)\),
\(\E_{\rm fp}[F]=\loopmax/\sum_{r=1}^{\loopmax}F(r)^{-1}\). Uniform sampling gives the arithmetic mean instead. Since the harmonic mean
is smaller than the arithmetic mean for $\rho>0$,
$\E_{\rm unif}[F]/\E_{\rm fp}[F]>1$.

\paragraph{(c) Noise-proportional.}
\label{sec:method-noiseprop}
We also study a noise-proportional distribution to test the hypothesis
that noisier inputs benefit from more loop iterations, since their
observed tokens provide less direct information about the clean
sequence.
We split \(\{1,\ldots,\loopmax\}\) into \(G\) consecutive groups
\(\mathcal S_1,\ldots,\mathcal S_G\) of equal size, from small to large
loop counts. Using antithetic time sampling, we sort the training
microbatches into \(G\) equal bands of increasing noise, and band \(b\) draws
\(\loops\) uniformly from \(\mathcal S_b\)
(Figure~\ref{fig:flopprop-budget}b,c, Appendix~\ref{app:loop-sampling}).
The marginal distribution of loop counts remains uniform,
so this distribution matches uniform sampling in expected training
FLOPs while isolating the effect of this particular
coupling between noise and loop count.

\begin{figure*}[t]
  \centering
  \includegraphics[width=\textwidth]
    {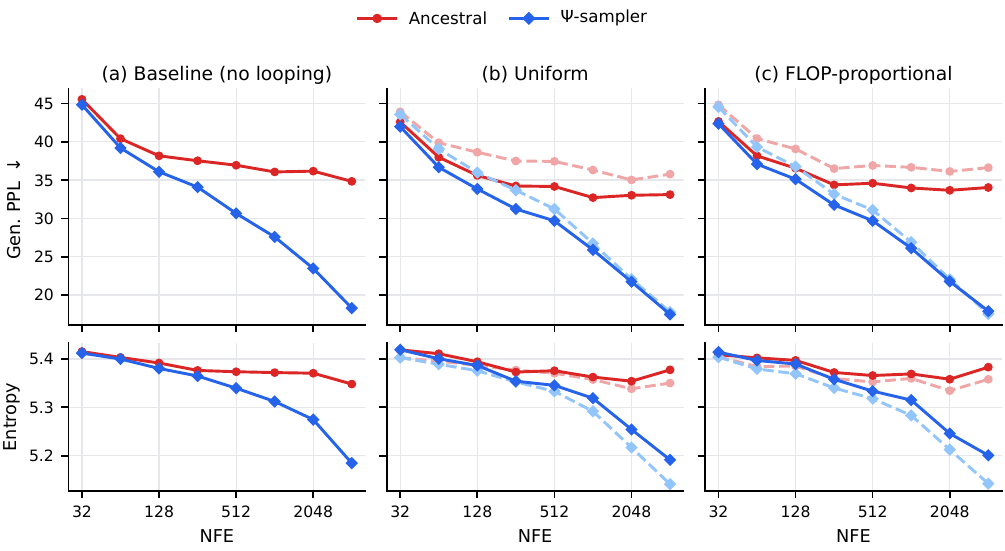}
  \caption{\textbf{Generation quality and diversity.} Generative perplexity (top) and unigram entropy (bottom) against NFE. Dashed lines are $\loops=1$ and solid lines are $\loops=6$. Looping lowers generative perplexity at the same NFE without lowering entropy.}
  \label{fig:reference-ppl-entropy}
\end{figure*}

\section{Experimental Setup}
\label{sec:setup}

\paragraph{Model and datasets.}
All models use the \(170\)M-parameter \methodpp{} \texttt{small} DiT
(\(12\) blocks, width \(768\), sequence length \(1{,}024\), GPT-2 tokenizer)
with its efficient curriculum. Looped models replay blocks \(1\)--\(2\) with
\(\loopmax=12\), a choice justified by the ablation in
Figure~\ref{fig:loop-layer-ablation}a. We train on two datasets:
OpenWebText (OWT)~\citep{Gokaslan2019OpenWeb} and the \texttt{sample-10BT} subset of
FineWeb-Edu~\citep{penedo2024fineweb}. On both, the final \(100\)k documents are held out for validation.
Following \methodpp{}~\citep{deschenaux2026diffusion}, the
non-looped baseline is trained for \(1\)M optimizer steps with a batch size
of \(512\).
We train with sequence packing following \citet{raffel2020exploring}, and
Appendix~\ref{app:training-config} gives more details.

\paragraph{Iso-FLOP budgets.}
We compare runs at the same training FLOPs. After a given number of steps, a
run has used $\text{step}\times B\times\Floop$ FLOPs, where $B=512$
sequences and $\Floop=\Ftrain(\Es)$ is the training cost per sequence at the
run's mean loop count (Appendix~\ref{app:flops}). The cost relative to the
baseline, $\Floop/\Fbase$, depends only on the loop-count distribution
(Table~\ref{tab:flopprop}). We use the Chinchilla convention, and the Kaplan
variant only rescales the axis (Figure~\ref{fig:kaplan-elbows}). The FLOPs of the
baseline's 1M steps buy \(655{,}420\) steps for the uniform run and
\(688{,}915\) for the FLOP-proportional run. On average, a training step runs
\(6.50\) loop iterations with uniform or noise-proportional sampling, and
\(5.72\) with FLOP-proportional sampling. Every \(5\)k optimizer steps, we
record validation NLL at \(\loops\in\{1,6,8,12\}\). The noise-proportional
run is also scored with each noise band at a low, middle, or high loop count
from the band it was trained on (Appendix~\ref{app:loop-sampling}). None of
its evaluations reaches the other two frontiers (Figure~\ref{fig:loop-layer-ablation}b).

\paragraph{Looping at inference.}
At inference the loop count is a free parameter. We run checkpoints at
\(\loops=1,6,8,12\) and also with \emph{adaptive
looping}~\citep{lee2026looped}, which picks \(\loops\) separately for each
sequence. A sequence keeps looping until its hidden state barely moves, i.e., until
\(\lVert\mathbf h^{(r)}-\mathbf h^{(r-1)}\rVert_F/\lVert\mathbf h^{(r)}\rVert_F<\epsilon=0.1\),
and never goes past \(\loopmax=12\).

\paragraph{Evaluation.}
All inference results use OWT checkpoints, evaluated along three axes:
(a) generative perplexity (Gen.\ PPL) under GPT-2 Large and unigram entropy
of generated samples, across numbers of function evaluations (NFE, i.e.,
denoiser calls) (\S\ref{sec:fixed-loop-results}), (b) zero-shot perplexity
on seven out-of-domain datasets (\S\ref{sec:zero-shot-ppl}), and
(c) downstream accuracy on ten multiple-choice tasks
(\S\ref{sec:flopprop-inference}).
Appendices~\ref{app:generation}--\ref{app:zero-shot-mcq} give the full
protocols.

\begin{table}[t]
\centering
\small
\setlength{\tabcolsep}{3.5pt}
\renewcommand{\arraystretch}{0.95}
\resizebox{0.9\textwidth}{!}{%
\begin{tabular}{lrrrrrrrrr}
\toprule
& & \multicolumn{6}{c}{Fixed looping} & \multicolumn{2}{c}{Adaptive looping} \\
\cmidrule(lr){3-8}\cmidrule(lr){9-10}
& & \multicolumn{3}{c}{Uniform} & \multicolumn{3}{c}{FLOP-prop.} & & \\
\cmidrule(lr){3-5}\cmidrule(lr){6-8}
Dataset & Baseline & \(S{=}6\) & \(S{=}8\) & \(S{=}12\) & \(S{=}6\) & \(S{=}8\) & \(S{=}12\) & Uniform & FLOP-prop. \\
\midrule
PTB              & 94.86 & 85.18 & 84.01 & \textbf{83.97} & 85.47 & 85.19 & 85.55 & 85.14 & 85.43 \\
WikiText-103     & 34.46 & 31.70 & 31.39 & \textbf{31.31} & 31.65 & 31.46 & 31.44 & 31.75 & 31.71 \\
LM1B             & 74.21 & 71.16 & 70.59 & 70.38 & 70.35 & \textbf{70.05} & 70.15 & 71.22 & 70.47 \\
LAMBADA          & 51.83 & 48.02 & 47.76 & 47.70 & 46.38 & \textbf{46.33} & 46.45 & 48.21 & 46.56 \\
AG News          & 69.95 & 59.21 & 57.93 & \textbf{57.31} & 59.88 & 58.77 & 58.35 & 58.68 & 59.63 \\
PubMed           & 44.91 & 38.91 & 38.30 & \textbf{37.95} & 39.16 & 38.52 & 38.11 & 38.82 & 39.23 \\
arXiv            & 39.78 & 35.02 & 34.47 & \textbf{34.13} & 35.27 & 34.71 & 34.37 & 34.97 & 35.31 \\
\midrule
\textbf{Average} & 58.57 & 52.74 & 52.06 & \textbf{51.82} & 52.59 & 52.15 & 52.06 & 52.68 & 52.62 \\
\bottomrule
\end{tabular}}
\caption{\textbf{Zero-shot perplexity} ($\downarrow$). Mean over five seeds. Adaptive looping uses $\epsilon=0.1$ and $S_{\max}=12$. Bold marks the best value in each row. Appendix~\ref{app:zero-shot-ppl} gives $S=1$ results and standard deviations.}
\label{tab:zero-shot-ppl-main}
\end{table}

\section{Results}
\label{sec:results}

\paragraph{Where to loop.}
On OpenWebText (\(50\)k steps, one seed), two-block slices starting anywhere
in blocks \(1\)--\(5\) perform about equally, late slices with a short tail
are clearly worse, and raising \(\loopmax\) beyond \(12\) gives no further
gain (Figure~\ref{fig:loop-layer-ablation}a,
Appendix~\ref{app:loop-ablation}). All looped models therefore loop blocks
\(1\)--\(2\) with \(\loopmax=12\).

\subsection{Looping is training-compute efficient}
\label{sec:isoflop-training}

\begin{figure*}[t]
  \centering
  \includegraphics[width=\textwidth]
    {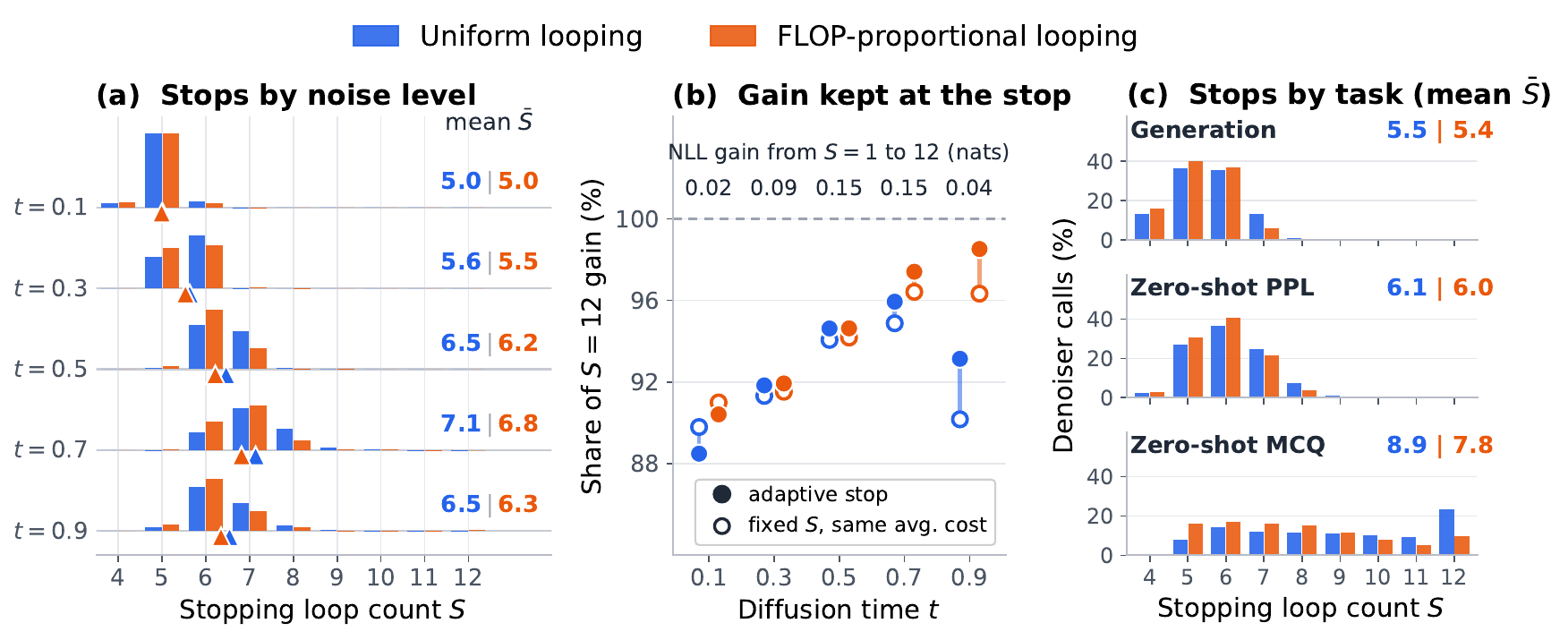}
  \caption{\textbf{Where adaptive looping stops} ($\epsilon=0.1$, $\loopmax=12$). (a)~Stopping loop count of single denoiser calls on the eight datasets at each diffusion time, with triangles at the mean. (b)~Share of the NLL gain from $\loops=1$ to $12$ kept at the adaptive stop (filled) and at a fixed loop count with the same average cost (open). The top row gives the total gain of the uniform checkpoint. (c)~Stopping loop counts over all noise levels for each evaluation.}
  \label{fig:adaptive-stopping}
\end{figure*}

We compare each looped run with the non-looped baseline at matched training
FLOPs on both datasets (Figure~\ref{fig:training-elbow}, with per-step curves in
Figure~\ref{fig:training-curves}). Both looped runs stay below the baseline
at every budget, and the savings are larger on FineWeb-Edu than on
OpenWebText. Evaluating at \(\loops=6\), \(8\), or \(12\) gives nearly the
same curve, so a few loop iterations capture almost all of the gain.
FLOP-proportional sampling closely matches the final NLL of uniform
sampling (\(3.159\) vs.\ \(3.154\) on OpenWebText and \(3.122\) vs.\
\(3.119\) on FineWeb-Edu), and it is better at \(\loops=1\) (\(3.289\) vs.\
\(3.306\)), the loop count it trains on most often.

\findingbox{\textbf{Looping improves training efficiency in UDLMs.}
Across OpenWebText and FineWeb-Edu, both looped models reach the
non-looped baseline's validation NLL with \(2.7\)--\(3.3\times\)
fewer training FLOPs.}

\label{sec:noiseprop-results}
Noise-proportional sampling has the same per-step cost as uniform sampling,
so the two runs are iso-FLOP at every step. We stopped it after \(16\) epochs
of OpenWebText. Besides fixed loop counts, we score it with band-consistent
evaluation, which scores each noise band only at loop counts it was trained
on (Appendix~\ref{app:loop-sampling}). None of these six evaluations reaches
the uniform or FLOP-proportional frontier
(Figure~\ref{fig:loop-layer-ablation}b), and the other two distributions
reach its best NLL with roughly \(30\%\) less compute.

\findingbox{\textbf{The noise-proportional distribution offers no benefit.}
At matched OpenWebText training FLOPs, assigning deeper loops to noisier
inputs underperforms uniform and FLOP-proportional sampling
(the best result trails FLOP-proportional by \(+0.029\) NLL).
This is consistent with Figure~\ref{fig:loop-reliability}b, where the
loop gain at inference peaks at intermediate diffusion times rather than
at the noisiest inputs.}

\subsection{Looping at inference improves generation}
\label{sec:fixed-loop-results}

We sample from each checkpoint with the ancestral and \(\Psi\) samplers
across NFE budgets and compare loop counts at matched NFE
(Figure~\ref{fig:reference-ppl-entropy}, Appendix~\ref{app:generation}).
Looping lowers Gen.\ PPL without reducing entropy. With ancestral sampling
the gain is small at the lowest NFE and largest at \(128\)--\(1{,}024\) NFE. With
the \(\Psi\)-sampler it is smaller and shrinks at large NFE, where the
sampler alone already gives low Gen.\ PPL. These comparisons fix NFE, not
FLOPs: \(\loops=6\) costs \(1.48\times\) the inference FLOPs of \(\loops=1\)
and \(\loops=12\) costs \(2.05\times\), while \(\loops=6\) or \(8\) already
gives most of the gain. Adaptive looping averages about \(5.5\) loop
iterations, slightly cheaper than fixed \(\loops=6\), and matches or beats
it at most sampler--NFE points.

\findingbox{\textbf{Looping at inference matters more than maximizing depth.}
At matched NFE, looped models reach up to \(9.8\%\) lower Gen.\ PPL than
the non-looped baseline, with most gains by \(\loops=6\)--\(8\). Adaptive looping retains
comparable quality with fewer loop iterations.}

\subsection{Looping at inference improves zero-shot likelihood}
\label{sec:zero-shot-ppl}

\begin{figure*}[t]
  \centering
  \includegraphics[width=\textwidth]{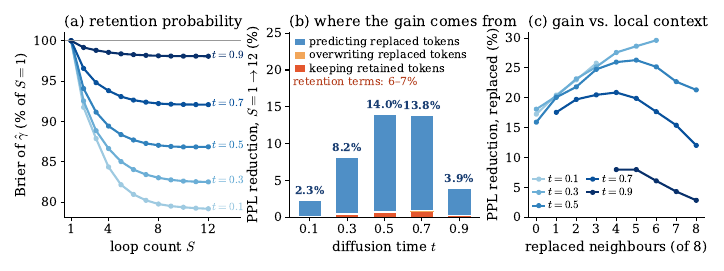}
  \caption{\textbf{Where the loop gain comes from.} Single denoiser calls of the uniform checkpoint, averaged over eight datasets. (a)~Brier score of the retention probability $\hat\gamma_t^\ell$ (Equation~\ref{eqn:udm-reliability}) relative to $\loops=1$. (b)~Perplexity reduction from $\loops=1$ to $12$, split into the terms of Equation~\ref{eqn:reliability-decomposition}. (c)~Reduction at replaced tokens against the number of replaced neighbours within four positions.}
  \label{fig:loop-reliability}
\end{figure*}

We bound the perplexity of each checkpoint on seven out-of-domain datasets
(Table~\ref{tab:zero-shot-ppl-main}). Run at \(\loops=1\), both looped
checkpoints are worse than the non-looped baseline on every dataset
(Table~\ref{tab:zero-shot-ppl-all}), so loop training alone does not give
the gain. Looping at inference reverses this: every fixed \(\loops>1\) and
adaptive looping beat the baseline on every dataset. Unlike generation, likelihood keeps
improving with depth, and \(\loops=12\) is the best loop count of the
uniform checkpoint on every dataset.

\paragraph{Adaptive looping stops where looping stops helping.}
On single denoiser calls (Figure~\ref{fig:adaptive-stopping}a), both
checkpoints stop early at low and high noise and loop longest at
intermediate diffusion times. This tracks where looping helps: the NLL gain
from \(\loops=1\) to \(12\) is several times larger at \(t=0.5\)--\(0.7\)
than at \(t=0.1\) or \(0.9\). At the stop, adaptive looping keeps
\(88\%\)--\(98\%\) of the \(\loops=12\) gain and matches or beats a fixed
loop count with the same average cost at every \(t\ge0.3\)
(Figure~\ref{fig:adaptive-stopping}b). Generation stops earliest and MCQ
scoring latest (Figure~\ref{fig:adaptive-stopping}c). On zero-shot
likelihood, adaptive looping stays close to \(\loops=6\) while skipping
\(10\%\)--\(28\%\) of loop iterations (Appendix~\ref{app:zero-shot-ppl}).

\paragraph{Looping mostly improves the prediction at replaced tokens.}
We apply the decomposition of \S\ref{sec:method-reliability} to single
denoiser calls, scoring \(\widehat\x\) after each loop iteration
(Appendix~\ref{app:single-call}). The NLL splits into three terms: keeping
retained tokens, overwriting replaced tokens, and predicting the clean token
at replaced positions. Which tokens to keep is largely settled without
looping: at \(\loops=1\), \(\hat\gamma_t^\ell\) already separates retained
from replaced tokens (AUROC \(\ge0.96\)), and extra loop iterations improve
it further mainly at small \(t\) (Figure~\ref{fig:loop-reliability}a).
Instead, \(93\%\)--\(94\%\) of the perplexity reduction from \(\loops=1\) to
\(12\) comes from better contextual prediction at replaced tokens, peaking
at intermediate \(t\) (Figure~\ref{fig:loop-reliability}b). For
\(t\ge0.5\), this gain is largest when a replaced token's neighbourhood is
partly corrupted and smaller when nearly all neighbours are replaced
(Figure~\ref{fig:loop-reliability}c). The FLOP-proportional checkpoint
shows the same pattern.

\findingbox{\textbf{Out-of-domain likelihood needs looping at inference.}
Loop-trained checkpoints at \(\loops=1\) underperform the non-looped
baseline, whereas looping them at inference improves the likelihood bound.
Most of this gain comes from better contextual prediction at replaced
tokens.}

\subsection{Looping improves zero-shot accuracy}
\label{sec:flopprop-inference}

\begin{figure*}[t]
  \centering
  \includegraphics[width=\textwidth]
    {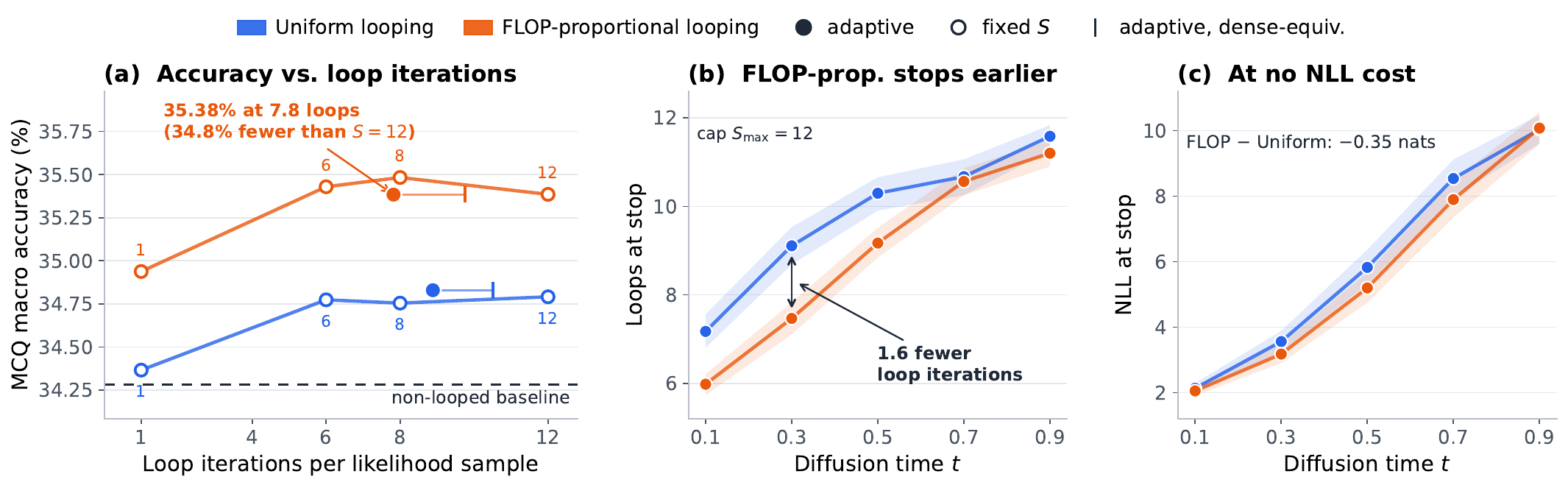}
  \caption{\textbf{Adaptive looping on multiple-choice tasks.} (a)~Average accuracy on the ten tasks against loop iterations, over five seeds. Ticks show the dense-equivalent cost of adaptive looping. (b, c)~Loop count and NLL at the adaptive stop on 64 sequences, with 95\% bands. FLOP-proportional stops earlier at low and middle noise levels with no higher NLL.}
  \label{fig:zero-shot-mcq-frontier}
\end{figure*}

We evaluate on ten zero-shot multiple-choice tasks
(Figure~\ref{fig:zero-shot-mcq-frontier}a,
Appendix~\ref{app:zero-shot-mcq}). The FLOP-proportional checkpoint beats
the uniform one at every matched loop policy, and looping at inference
raises its accuracy above its own \(\loops=1\). With adaptive looping it
matches fixed \(\loops=12\) with about a third fewer loop iterations. Gains are largest on HellaSwag,
OpenBookQA, PIQA, and BoolQ and negligible on MathQA and WinoGrande. The
FLOP-proportional checkpoint also beats the uniform one on FineWeb-Edu
(Table~\ref{tab:fineweb-edu-mcq}).

\paragraph{The FLOP-proportional checkpoint stops looping earlier.}
FLOP-proportional sampling puts more weight on small loop counts
(\S\ref{sec:method-loop}), so its slice may need fewer iterations. Under
adaptive looping on MCQ, it runs \(12\%\) fewer loop iterations than the
uniform checkpoint (\(7.82\) vs.\ \(8.89\),
Table~\ref{tab:zero-shot-mcq-adaptive}) while being more accurate. On
matched inputs, it also reaches a lower NLL at the stopping point
(Figure~\ref{fig:zero-shot-mcq-frontier}b,c). The saving is concentrated at
low and intermediate noise and disappears on the noisiest inputs, which
still get deep computation. Generation and zero-shot likelihood show the
same, smaller shift (Figure~\ref{fig:adaptive-stopping}c).

The trade-off is sample quality. At matched loop counts, the
FLOP-proportional checkpoint has a higher Gen.\ PPL than the uniform one at
most points, with nearly unchanged entropy, and the gap widens with more
loop iterations (Table~\ref{tab:generation-ppl}).

\findingbox{\textbf{Depth allocation produces task-dependent trade-offs.}
FLOP-proportional training improves zero-shot MCQ accuracy, and the
resulting checkpoint stops earlier under adaptive looping, reaching \(35.38\%\)
macro accuracy, comparable to fixed \(\loops=12\) with \(34.8\%\)
fewer loop iterations. However, its Gen.\ PPL is \(1.55\%\) higher on
average than that of the uniform checkpoint at matched loop counts.}

\section{Related Work}
\label{sec:related}

\paragraph{Discrete diffusion language models.}
Discrete diffusion models use structured categorical corruption
\citep{austin2021structured,hoogeboom2021argmax}, with much of the
literature focusing on masked diffusion
\citep{sahoo2024simple,shi2025simplified,ou2024RADD,lou2024discrete,nie2025large}.
UDLMs instead corrupt tokens into ordinary vocabulary tokens,
allowing previously generated tokens to be revised
\citep{schiff2025simple,sahoo2025diffusionduality,sahoo2026scaling}.
Existing acceleration methods largely reduce the number of
denoising steps through improved sampling or distillation
\citep{song2023consistency,kim2024consistency,hu2026cmt,peng2025path,
israel2025accelerating,luxembourg2025plan,dang2025inference}.
Our work instead studies recurrent computation within each
denoising step, leaving the diffusion objective and sampler unchanged.

\paragraph{Looped transformers.}
Shared-weight recurrence increases effective depth without adding
parameters, building on Universal Transformers, equilibrium models,
and parameter-shared architectures
\citep{dehghani2018universal,bai2019deep,lan2019albert,
giannou2023looped}.
Recent work studies iterative computation, test-time scaling,
and the quality--compute trade-offs of recurrent depth
\citep{saunshi2025reasoning,geiping2025scaling,
zhu2025scalinglatent,mcleish2025teaching,prairie2026parcae,
schwethelm2026much}.
We investigate these questions in UDLM denoisers, including how
recurrent depth should be sampled during training.

\paragraph{Looped diffusion denoisers.}
Closest to our work, \citet{lee2026looped} study recurrent depth
in masked diffusion, while related approaches explore recursive
blocks and fixed-point formulations
\citep{carballocastro2026recursive,
miele2026fixedpointmaskedgenerativemodeling}.
In standard absorbing masked diffusion, revealed tokens remain
fixed, whereas UDLMs must denoise observations of uncertain
reliability and can revise any token between diffusion steps.
We study whether recurrence remains effective under this
different corruption structure, and how its computation
should be allocated.

\paragraph{Recurrence across denoising steps.}
Self-conditioning reuses clean-token predictions across
diffusion steps
\citep{chen2023analog,strudel2022self,dieleman2022continuous}.
Related approaches propagate latent states across steps or
introduce additional denoising recursion
\citep{jo2026loopholing,suleymanzade2026thinkingloopedflows,
yoo2026self,cameron2026stepforwardksteps}.
Our approach instead loops a slice of transformer blocks \emph{within}
each denoiser evaluation, updating hidden representations
while keeping the noisy sequence and diffusion time fixed.
These directions are complementary.

\section{Discussion and Conclusion}

We studied the effect of looping in UDLMs by repeating a two-block slice of
the denoiser and training it with random loop counts. Looped models reach
the validation NLL of the non-looped baseline with up to $2.8\times$ fewer
training FLOPs on OpenWebText and $3.3\times$ fewer on FineWeb-Edu. We show
that at inference, looping lowers generative perplexity by up to $9.8\%$ at the same
NFE and zero-shot perplexity by $11.5\%$ on average. With adaptive looping,
the model needs $26$--$35\%$ fewer loop iterations on multiple-choice tasks
and keeps the same accuracy. Training on lower loop counts more often helps
further: on the same tasks, the model needs $12\%$ fewer loop iterations
than with uniform sampling, without degrading accuracy. We plan to extend
this study by comparing how looping helps masked and uniform diffusion,
especially on reasoning tasks.


\subsubsection*{Acknowledgments}
We thank Sanghyun Lee and Chunsan Hong for helpful correspondence and for sharing details of the
experimental setup of \citet{lee2026looped}. We are grateful to AWS and Modal
for generously providing compute through their research grants.

\clearpage

\bibliography{colm2026_conference}
\bibliographystyle{colm2026_conference}

\clearpage
\appendix
\setcounter{topnumber}{4}
\setcounter{totalnumber}{6}
\renewcommand{\topfraction}{0.95}
\renewcommand{\textfraction}{0.05}
\renewcommand{\floatpagefraction}{0.85}
\etocdepthtag.toc{appendix}
\etocsettagdepth{main}{none}
\etocsettagdepth{appendix}{subsection}
\etocsettocstyle{\section*{Appendix Contents}\vspace{1ex}}{\clearpage}
\tableofcontents

\section{Technical primer: discrete diffusion, UDLMs, \method{}, and \methodpp{}}
\label{app:primer}

This section provides more details about DUO/DUO++ and their training objectives summarized in \S\ref{sec:background}.

\subsection{Notation}
\label{app:primer-notation}
\(\mathbf e_i\in\R^K\) is the \(i\)-th one-hot vector,
\(\onehotset=\{\mathbf e_1,\ldots,\mathbf e_K\}\) is the set of one-hot
vectors, and \(\Delta^K\) is the \(K\)-class simplex. Superscripts \(\ell\)
index positions and subscripts \(i,j\) index vocabulary entries. Time runs
from clean data (\(t=0\)) to noise (\(t=1\)), and \(s<t\) is the less noisy
time. \(\alpha_t\) is the nonincreasing noise schedule. \(\vone\) is the
all-ones vector, \(\mathbf u=\vone/K\) is the uniform distribution, \(\odot\)
is the elementwise product, and matrices act on column probability vectors.
\(\nelbo\) is the negative evidence lower bound. 

\subsection{Sequence model and independence assumptions}
\label{app:primer-assumptions}
Clean tokens are not assumed independent. Corruption is independent across
positions given the clean sequence, and the reverse model factorizes given
the noisy sequence:
\begin{equation}
\label{eqn:primer-sequence-assumptions}
q_t(\ztL\mid\xL)
 =\prod_{\ell=1}^{L}q_t(\z_t^\ell\mid\x^\ell),
\qquad
p_\theta(\z_s^{1:L}\mid\ztL)
 =\prod_{\ell=1}^{L}
 p_\theta^\ell(\z_s^\ell\mid\ztL).
\end{equation}
Each reverse factor attends to the whole of \(\ztL\), so only the tokens
sampled within one reverse step are independent. The forward process is
Markov per position, \(q(\z_t^\ell\mid\z_s^\ell,\x^\ell)=q(\z_t^\ell\mid\z_s^\ell)\),
and we assume \(\alpha_t\) is differentiable and all probabilities inside
logarithms are positive.

\paragraph{Discrete diffusion language models.}
A D3PM~\citep{austin2021structured} uses a column-stochastic transition
\(\bm Q_{t\mid s}\), where \([\bm Q_{t\mid s}]_{ij}\) is the probability of
moving from token \(j\) to \(i\), and a cumulative kernel
\(\bar{\bm Q}_t=\bm Q_{t\mid s}\bar{\bm Q}_s\) with
\(q_t(\z_t\mid\x)=\cat(\z_t;\bar{\bm Q}_t\x)\). Bayes' rule gives
\begin{equation}
\label{eqn:primer-d3pm-posterior-overview}
q(\z_s\mid\z_t,\x)
=\cat\!\left(
\z_s;\
\frac{(\bm Q_{t\mid s}^{\!\top}\z_t)\odot(\bar{\bm Q}_s\x)}
{\z_t^{\!\top}\bar{\bm Q}_t\x}
\right).
\end{equation}
In the clean-token (\(x_0\)) parameterization the reverse kernel substitutes
the denoiser output for the unknown clean token,
\(p_\theta^\ell(\z_s^\ell\mid\ztL)
:=q(\z_s^\ell\mid\z_t^\ell,\x^\ell\leftarrow\denoise^\ell(\ztL,t))\).
Summing the resulting posterior KLs over positions and times gives the
diffusion term of the NELBO. An absorbing \(\bar{\bm Q}_t\) gives MDLMs~\citep{sahoo2024simple}.

\paragraph{Uniform diffusion language models.}
UDLMs~\citep{schiff2025simple,lou2024discrete} use
\begin{equation}
\label{eqn:primer-usdm-kernels-overview}
q_t(\z_t\mid\x)=\cat(\z_t;\at\x+(1-\at)\mathbf u),
\qquad
\bm Q_{t\mid s}
=\rho_{t\mid s}\bm I+(1-\rho_{t\mid s})\mathbf u\vone^\top,
\qquad
\rho_{t\mid s}=\frac{\at}{\as}.
\end{equation}
Substituting into Equation~\ref{eqn:primer-d3pm-posterior-overview},
\begin{equation}
\label{eqn:primer-usdm-posterior-overview}
q(\z_s\mid\z_t,\x)
=\cat\!\left(
\z_s;\
\frac{
[\rho_{t\mid s}\z_t+(1-\rho_{t\mid s})\mathbf u]
\odot[\as\x+(1-\as)\mathbf u]}
{\z_t^\top[\at\x+(1-\at)\mathbf u]}
\right).
\end{equation}

\paragraph{Continuous-time loss.}
As the number of steps grows, the prior and reconstruction terms vanish and
the loss becomes the rate
\(f_t=\lim_{\delta\downarrow0}\delta^{-1}
\KL(q_{t-\delta\mid t}\|p_{\theta,t-\delta\mid t})\)~\citep{schiff2025simple}.
Given \(\z_t=\mathbf e_i\), let
\[
\bar{\x}=K\at\x+(1-\at)\vone,\qquad
\bar{\x}_\theta=K\at\denoise(\ztL,t)+(1-\at)\vone,
\qquad
r_j=\frac{\bar x_j}{\bar x_i},\quad
r_{\theta,j}=\frac{\bar x_{\theta,j}}{\bar x_{\theta,i}} .
\]
Then
\begin{equation}
\label{eqn:primer-usdm-loss-overview}
f_t(\z_t,\denoise;\x)
=-\frac{\dot\alpha_t}{K\at}
\sum_{j\ne i}
\left[
r_{\theta,j}-r_j+r_j\log\frac{r_j}{r_{\theta,j}}
\right]\ge 0,
\qquad
\nelbo=\E_{\xL,t,\ztL}\sum_{\ell=1}^{L}f_t^\ell ,
\end{equation}
where \(\dot\alpha_t=d\alpha_t/dt\le0\). The \(f_{\mathrm{Duo}}\) of
Equation~\ref{eqn:nelbo} is an implementation of this same integrand.

\paragraph{The Diffusion Duality and \method{}.}
\method{} draws a Gaussian latent \(\w_t=\atg\x+\sqrt{1-\atg^2}\,\beps\in\R^K\)
with signal coefficient \(\atg\) and
\(\beps\sim\mathcal N(\mathbf0,\bm I)\), and sets \(\z_t=G(\w_t)\), where
\(G(\w)=\mathbf e_{\arg\max_i w_i}\) returns the largest coordinate of \(\w\)
as a one-hot vector. With
\(\phi,\Phi\) the standard-normal density and CDF and
\(\nu(a)=a/\sqrt{1-a^2}\),
\begin{equation}
\label{eqn:primer-duo-map-overview}
p_\star(a)=\int_{\R}\phi(z-\nu(a))\Phi(z)^{K-1}\,dz,
\qquad
\Tmap(a)=\frac{Kp_\star(a)-1}{K-1},
\end{equation}
where \(p_\star(a)\) is the probability that the shifted clean coordinate
beats all \(K-1\) competitors. Then~\citep{sahoo2025diffusionduality}
\begin{equation}
\label{eqn:primer-duo-pushforward-overview}
G_{\star}\mathcal N\!\left(\atg\x,(1-\atg^2)\bm I\right)
=\cat\!\left(\cdot;\
\Tmap(\atg)\x+[1-\Tmap(\atg)]\mathbf u\right),
\end{equation}
where \(G_\star P\) is the distribution of \(G(W)\) for \(W\sim P\). Hence
any statistic of the hard state has the same expectation under the
Gaussian draw and under the UDLM marginal at \(\at=\Tmap(\atg)\). This holds
for one-time marginals only, not for distributions over full trajectories. With
\(\mathbf y_\tau(\w)=\softmax(\w/\tau)\) for a temperature \(\tau>0\), the curriculum objective is
\begin{equation}
\label{eqn:primer-duo-curriculum-overview}
\mathcal L_{\mathrm{Duo}}^{(\tau)}
=\E_{\xL,t,\w_t^{1:L}}
\sum_{\ell=1}^{L}
f_{\mathrm{Duo}}\!\left(
G(\w_t^\ell),
\denoise^\ell(\mathbf y_\tau(\w_t^{1:L}),t),
\Tmap(\atg);\x^\ell\right).
\end{equation}
The target stays the hard state, and only the denoiser input is softened. As
\(\tau\downarrow0\), \(\mathbf y_\tau(\w)\to G(\w)\) almost surely, and the
objective is the NELBO only in this limit with \(t\) sampled over all of
\([0,1]\). Otherwise it is a surrogate.

\paragraph{\methodpp{}: top-\(k\) curriculum.}
\methodpp{}~\citep{deschenaux2026diffusion} samples only the largest \(k\)
of the \(K-1\) unshifted competitors. For independent
\(V_r\sim\mathcal U(0,1)\),
\begin{equation}
\label{eqn:primer-duopp-topk-overview}
U^{(0)}=1,\qquad
U^{(r)}=U^{(r-1)}V_r^{1/(K-r)},\qquad
C_r=\sqrt{1-\atg^2}\,\Phi^{-1}(U^{(r)}),
\quad r=1,\ldots,k ,
\end{equation}
which gives descending Gaussian order statistics. The shifted clean
coordinate is sampled separately and inserted at its rank if it exceeds
\(C_k\). The denoiser input mixes only the \(k\) retained embeddings. In
the softmax normalizer, the clean coordinate is added exactly if it is not
retained, and each remaining zero-mean coordinate (\(K-k-1\) or \(K-k\) of
them) is replaced by the truncated moment
\[
\E[e^{X/\tau}\mid X<C_k]
=\exp\!\left(\frac{\sigma^2}{2\tau^2}\right)
\frac{\Phi((C_k-\sigma^2/\tau)/\sigma)}{\Phi(C_k/\sigma)},
\qquad X\sim\mathcal N(0,\sigma^2),\quad\sigma=\sqrt{1-\atg^2}.
\]
With a polynomial approximation of \(\Tmap\), the curriculum costs \(O(k)\)
rather than \(O(K)\) per token.

\paragraph{\(\Psi\)-sampling.}
With \(q_s(\z_s\mid\x)\) the earlier-time marginal, the clean-conditioned
\(\Psi\)-kernel~\citep{deschenaux2026diffusion} and its marginal property
are
\begin{equation}
\label{eqn:primer-psi-overview}
\begin{gathered}
\Psi_{s\mid t}(\z_s\mid\z_t,\x)
=\kappa_t q_{s\mid t}(\z_s\mid\z_t,\x)
 +(1-\kappa_t)q_s(\z_s\mid\x),\\
\sum_{\z_t}q_t(\z_t\mid\x)\Psi_{s\mid t}(\z_s\mid\z_t,\x)
=q_s(\z_s\mid\x),
\end{gathered}
\end{equation}
where \(\kappa_t\in[0,1]\) is a mixing weight. \(\kappa_t=1\) is ancestral sampling and \(\kappa_t=0\) redraws from the
earlier marginal. The learned sampler replaces \(\x\) by the prediction:
\begin{equation}
\label{eqn:primer-duopp-learned-psi}
\begin{aligned}
\widehat{\x}^{\,\ell}
&:=\denoise^\ell(\ztL,t),
&r_\theta^\ell
&:=q_{0\mid t}\!\left(\cdot\mid\z_t^\ell,
  \x^\ell\leftarrow\widehat{\x}^{\,\ell}\right),\\
p_{\mathrm{pred}}^\ell
&:=q_{s\mid t}\!\left(\cdot\mid\z_t^\ell,
  \x^\ell\leftarrow\widehat{\x}^{\,\ell}\right),
&p_{\mathrm{cor}}^\ell
&:=\as r_\theta^\ell+(1-\as)\mathbf u,\\
p_\Psi^\ell
&:=\kappa_t p_{\mathrm{pred}}^\ell
 +(1-\kappa_t)p_{\mathrm{cor}}^\ell,
&\z_s^\ell&\sim\cat(p_\Psi^\ell).
\end{aligned}
\end{equation}
It recovers the exact kernel only for a perfect prediction.

\subsection{Latent reliability under keep-or-replace corruption}
\label{app:reliability}

We derive Equation~\ref{eqn:reliability-decomposition} for the general
keep-or-replace kernel in Equation~\ref{eqn:general-corruption}.
Fix a time with \(0<\alpha_t<1\) and a position \(\ell\). The auxiliary
variable \(C_t^\ell\) records which branch was sampled at that time.
It does not identify an event along a continuous-time trajectory.
For an observed symbol \(\z_t^\ell=\mathbf e_i\), conditioning on the
other noisy positions and on \(t\) gives
\begin{align*}
\Pr(C_t^\ell=1,\z_t^\ell=\mathbf e_i\mid\z_t^{-\ell},t)
&=\alpha_t\pi_t^\ell(i),\\
\Pr(C_t^\ell=0,\z_t^\ell=\mathbf e_i\mid\z_t^{-\ell},t)
&=(1-\alpha_t)\nu(\mathbf e_i).
\end{align*}
Here \(\pi_t^\ell(i)\) is the contextual clean-token probability
from \S\ref{sec:method-reliability}. For a symbol outside the clean
vocabulary, such as \(\mask{}\), we set it to zero.
Normalizing these two probabilities gives \(\gamma_t^\ell\).
If the token was retained, the clean value equals the observation,
and if it was replaced, that observation is independent of the clean
value given the other noisy positions. Marginalizing over these
branches proves Equation~\ref{eqn:reliability-decomposition}.
For an ordinary observed token, it also gives
\[
\Pr(\x^\ell=\z_t^\ell\mid\ztL,t)
=\gamma_t^\ell+(1-\gamma_t^\ell)\pi_t^\ell(i),
\]
showing why retention and correctness are different events.
With absorbing masking, the retention probability is one for an
ordinary observed token and zero for the mask. With uniform
replacement, \(\nu(\mathbf e_i)=1/K\), yielding
Equation~\ref{eqn:udm-reliability}.

\paragraph{What the \methodpp{} denoiser estimates.}
Condition on \(\ztL\) with \(\z_t^\ell=\mathbf e_i\). The local loss in
Equation~\ref{eqn:primer-usdm-loss-overview} is a generalized KL divergence
between the ratios \(r_j\) and \(r_{\theta,j}\), and \(r_j\) enters it
linearly, so its expectation over the clean token is minimized at
\(r_{\theta,j}=\E[r_j\mid\ztL,t]\). Because \(\bar x_j/K\) is the probability
of observing \(\mathbf e_j\) from \(\x^\ell\), and \(\z_t^\ell\) depends on
the clean sequence only through \(\x^\ell\),
\[
\E[r_j\mid\ztL,t]
=\frac{\Pr(\z_t^\ell=\mathbf e_j\mid\z_t^{-\ell},t)}
      {\Pr(\z_t^\ell=\mathbf e_i\mid\z_t^{-\ell},t)}
=\frac{K\at\,\pi_t^\ell(j)+1-\at}{K\at\,\pi_t^\ell(i)+1-\at}.
\]
The predicted ratios match these for every \(j\) only if
\(K\at\denoise+(1-\at)\vone\) is proportional to
\(K\at\pi_t^\ell+(1-\at)\vone\). Both vectors sum to \(K\), so the minimizer
is \(\denoise=\pi_t^\ell\): the denoiser targets the contextual prediction,
not the clean-token posterior, and Equation~\ref{eqn:udm-reliability} turns
its output into the retention probability. In particular,
\(\bar x_{\theta,i}=(1-\at)/(1-\hat\gamma_t^\ell)\), which is \(K\) times the
normalizer of Equation~\ref{eqn:primer-usdm-posterior-overview} with \(\x\)
replaced by \(\denoise\). The argument assumes hard inputs. The soft inputs of
the \method{} curriculum carry additional confidence information, so there
the output need not equal \(\pi_t^\ell\).

\paragraph{Dependence across positions in UDLMs.}
The auxiliary branch indicators are sampled independently before
observing the noisy sequence, but their posterior need not factorize.
Let \(A\subseteq\{1,\ldots,L\}\) be the set of retained positions.
For UDLM corruption with \(0<\alpha_t<1\), the joint probability of this
retention pattern and the observation is
\(\alpha_t^{|A|}((1-\alpha_t)/K)^{L-|A|}q_{\mathrm{data}}(\x^A=\z_t^A)\),
where the data term is the probability that the clean tokens in \(A\)
match the observed values (one when \(A\) is empty). Dropping the factor
common to all subsets,
\[
\Pr\bigl(\{\ell:C_t^\ell=1\}=A\mid\ztL,t\bigr)
\propto
\left(\frac{K\alpha_t}{1-\alpha_t}\right)^{|A|}
q_{\mathrm{data}}(\x^A=\z_t^A).
\]
Dependencies among clean tokens can therefore couple posterior retention
judgments even though the forward corruption is independent across
positions given the clean sequence.

\section{Loop-count distributions}
\label{app:loop-sampling}

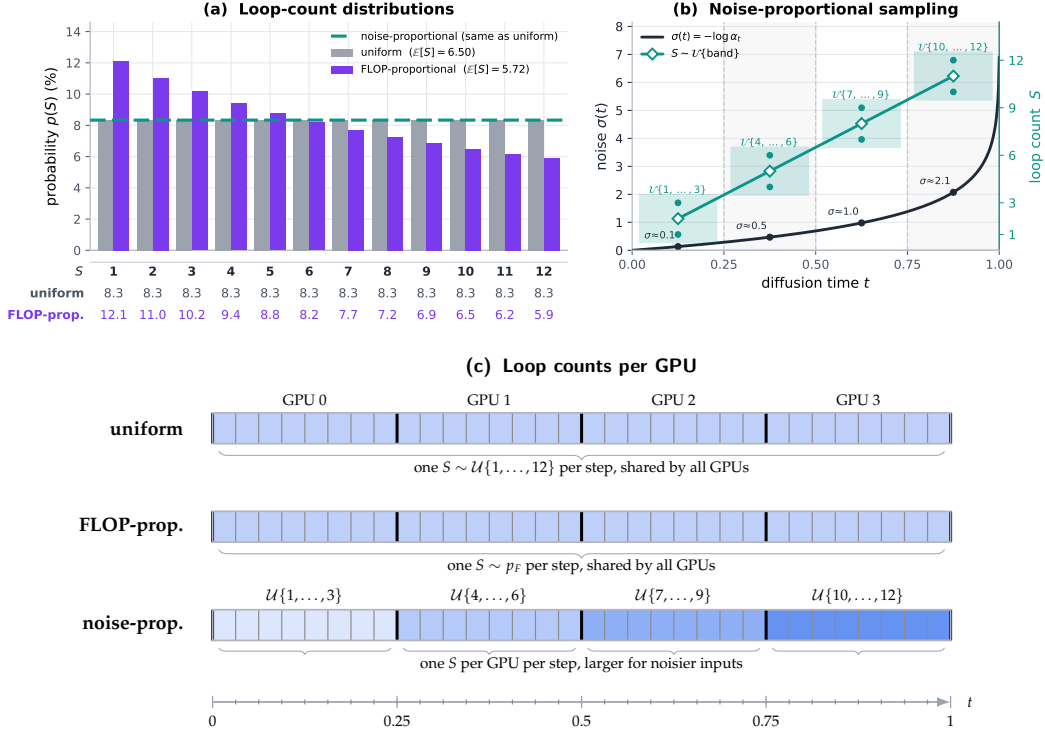
\begin{figure}[t]
\centering
\includegraphics[width=\textwidth]{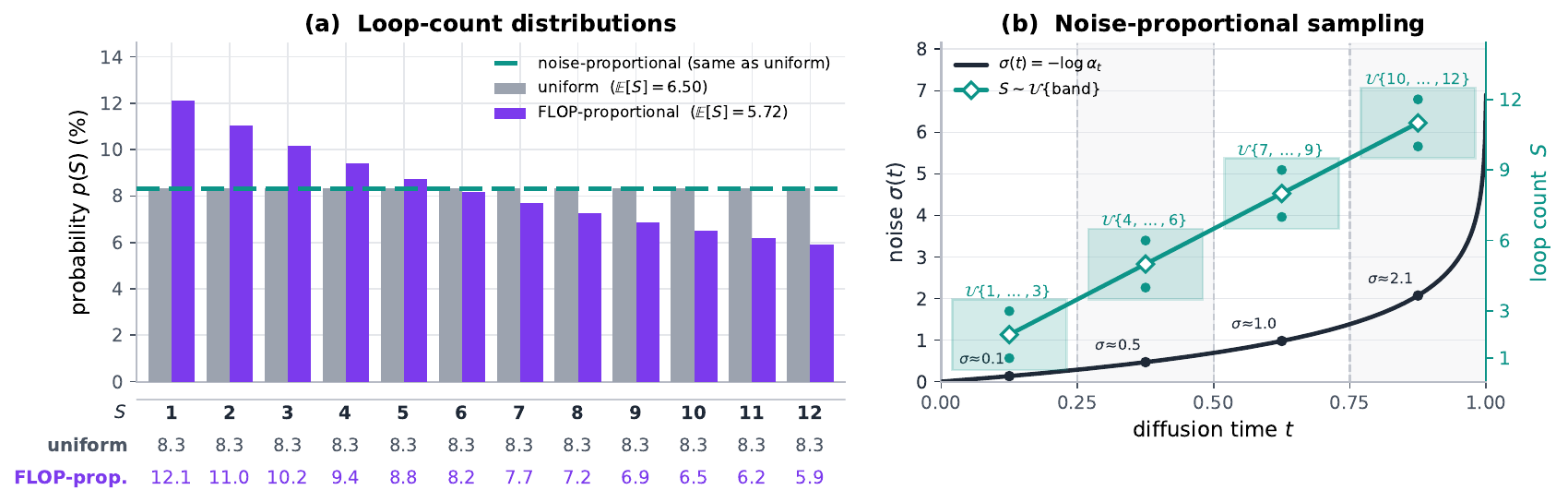}

\vspace{6pt}
\resizebox{0.86\textwidth}{!}{%
\begin{tikzpicture}[
  font=\small,
  gatick/.style={draw=black!45,line width=0.55pt},
  gpusep/.style={draw=black,line width=1.4pt},
  frame/.style={draw=plgray!80,line width=0.8pt},
  name/.style={anchor=east,font=\footnotesize\bfseries,text=black},
  slab/.style={font=\scriptsize,text=black},
  note/.style={font=\scriptsize,text=black},
]
\def\shardgrid#1{%
  \foreach \i in {1,...,31}{\draw[gatick] ({\i*0.375},#1) -- ({\i*0.375},{#1+0.5});}%
  \foreach \sx in {0,3,6,9,12}{\draw[gpusep] (\sx,#1) -- (\sx,{#1+0.5});}%
  \draw[frame] (0,#1) rectangle (12,{#1+0.5});%
}

\node[font=\small\sffamily\bfseries] at (6,6.15) {(c)\enspace Loop counts per GPU};
\foreach \d in {0,1,2,3}{\node[note] at (3*\d+1.5,5.62) {GPU \d};}

\fill[plblue!30] (0,4.9) rectangle (12,5.4);
\shardgrid{4.9}
\node[name] at (-0.35,5.15) {uniform};
\draw[decorate,decoration={brace,amplitude=4pt,mirror},plgray] (0.1,4.78) -- (11.9,4.78);
\node[slab] at (6,4.5) {one \(\loops\sim\mathcal{U}\{1,\dots,12\}\) per step, shared by all GPUs};

\fill[plblue!30] (0,3.3) rectangle (12,3.8);
\shardgrid{3.3}
\node[name] at (-0.35,3.55) {FLOP-prop.};
\draw[decorate,decoration={brace,amplitude=4pt,mirror},plgray] (0.1,3.18) -- (11.9,3.18);
\node[slab] at (6,2.9) {one \(\loops\sim p_{F}\) per step, shared by all GPUs};

\foreach \d/\sh/\lo/\hi in {0/16/1/3, 1/34/4/6, 2/52/7/9, 3/70/10/12}{
  \fill[plblue!\sh] (3*\d,1.7) rectangle (3*\d+3,2.2);
  \node[slab] at (3*\d+1.5,2.42) {\(\mathcal{U}\{\lo,\dots,\hi\}\)};
  \draw[decorate,decoration={brace,amplitude=3pt,mirror},plgray]
    (3*\d+0.1,1.58) -- (3*\d+2.9,1.58);
}
\shardgrid{1.7}
\node[name] at (-0.35,1.95) {noise-prop.};
\node[slab] at (6,1.3) {one \(\loops\) per GPU per step, larger for noisier inputs};

\draw[-{Latex[length=2.2mm]},plgray,line width=0.8pt] (0,0.7) -- (12,0.7);
\foreach \i in {1,...,31}{\draw[gatick] ({\i*0.375},0.66) -- ({\i*0.375},0.74);}
\foreach \tx/\tl in {0/0, 3/0.25, 6/0.5, 9/0.75, 12/1}{
  \draw[plgray,line width=0.6pt] (\tx,0.6) -- (\tx,0.8);
  \node[note,below] at (\tx,0.6) {\tl};
}
\node[note,anchor=west] at (12.15,0.7) {\(t\)};
\end{tikzpicture}%
}%
\caption{\textbf{Loop-count sampling.} (a) Loop-count distributions.
(b) Noise-proportional ranges. (c) Loop counts per GPU.}
\label{fig:flopprop-budget}
\end{figure}

Figure~\ref{fig:flopprop-budget} shows the three loop-count distributions of
\S\ref{sec:method-loop}. In uniform and FLOP-proportional training, all GPUs
share one loop count per optimizer step. The FLOP-proportional probabilities
depend only on $\rho$ (Equation~\ref{eqn:flopprop}).

In noise-proportional training, we draw the diffusion times of the global
batch antithetically and sort them before sharding. Each of the 32
micro-batches (4 GPUs $\times$ 8 accumulation steps) then gets a narrow range
of $t$ (Figure~\ref{fig:flopprop-budget}c). We group them into four bands.
Bands 1 to 4 sample $\loops$ from $\{1,2,3\}$, $\{4,5,6\}$, $\{7,8,9\}$, and
$\{10,11,12\}$. The average loop count stays at 6.5, as in uniform sampling.
At evaluation, we sort validation sequences into the same bands and use a loop
count from each band's training range.

\section{FLOP accounting convention}
\label{app:flops}

For the iso-FLOP comparisons in \S\ref{sec:setup}, we count a multiply-add as
two FLOPs and the backward pass as twice the forward pass, so one
training step costs $\Ftrain=3\Ffwd$.

\paragraph{Forward FLOPs.}
We extend the count of \citet[Appendix~F]{hoffmann2022training} to the DiT
denoiser \citep{peebles2023scalable}. The model has $n$ blocks of width $d$,
MLP width $4d$, AdaLN conditioning width $c$, timestep-frequency width $f$,
vocabulary $V$, and sequence length $L$. Table~\ref{tab:flop-components} lists
each term. With $\nexec$ executed blocks, the forward cost is
\begin{equation}
\label{eqn:fwd}
\Ffwd(\nexec) = \nexec\,F_{\mathrm{block}}
  + \underbrace{2(fc+c^2) + 4dc}_{\text{timestep MLP, final AdaLN}}
  + \underbrace{4LVd}_{\text{vocab terms}},
\end{equation}
where $F_{\mathrm{block}} = 24Ld^2 + 4L^2d + 12dc$ is the cost of one block.

For the non-looped \texttt{small} baseline ($\nexec=n$), this gives
$\Ffwd=370.71$ and $\Ftrain=1112.13$ GFLOP per sequence under Chinchilla, and
$\Ftrain=637.85$ under Kaplan, on both OWT and FineWeb-Edu. This is close to
the $6ND$ estimate of $1041.9$ GFLOP ($N=169.6$M, $D=L$), which skips the
attention scores.

\begin{table}[t]
\centering
\footnotesize
\begin{minipage}[t]{0.49\textwidth}
\centering
\setlength{\tabcolsep}{2.5pt}
\begin{tabular}[t]{@{}llc@{}}
\toprule
Component & Forward FLOPs & Counted \\
\midrule
input embedding\(^\dagger\) & \(2LVd\) & once \\
\(Q,K,V\) projections & \(6Ld^2\) & per block \\
\(QK^\top\), attn.\(\cdot V\) & \(4L^2d\) & per block \\
attention output & \(2Ld^2\) & per block \\
MLP (\(d_{\mathrm{ff}}=4d\)) & \(16Ld^2\) & per block \\
AdaLN modulation & \(12dc\) & per block \\
timestep MLP & \(2(fc+c^2)\) & once \\
final AdaLN & \(4dc\) & once \\
output logits\(^\dagger\) & \(2LdV\) & once \\
\bottomrule
\end{tabular}
\captionof{table}{\textbf{Forward FLOPs per sequence.} $^\dagger$Vocab terms, not counted under Kaplan.}
\label{tab:flop-components}
\end{minipage}\hfill
\begin{minipage}[t]{0.49\textwidth}
\centering
\setlength{\tabcolsep}{2.5pt}
\begin{tabular}[t]{@{}lcccc@{}}
\toprule
& & \multicolumn{2}{c}{\(\Floop/\Fbase\)} & \\
\cmidrule(lr){3-4}
Distribution & \(\Es\) & Chinchilla & Kaplan & Steps \\
\midrule
uniform     & \(6.500\) & \(1.526\times\) & \(1.917\times\) & \(655{,}420\) \\
FLOP-prop. & \(5.724\) & \(1.452\times\) & \(1.787\times\) & \(688{,}915\) \\
noise-prop. & \(6.500\) & \(1.526\times\) & \(1.917\times\) & \(655{,}420\)\(^\ast\) \\
\bottomrule
\end{tabular}
\captionof{table}{\textbf{Cost of looped training.} Mean loop count, cost per step relative to the baseline, and looped steps matching a 1M-step baseline under Chinchilla. $^\ast$Noise-proportional is trained on OWT only and stopped after about 273k steps.}
\label{tab:flopprop}
\end{minipage}
\end{table}

\paragraph{Looping.}
Looping runs $n_\midblock$ shared blocks $\loops$ times, so
$\nexec=n+(\loops-1)n_\midblock$ with the same number of parameters. Each extra
loop adds the cost of $n_\midblock$ blocks, so a training step at loop count
$\loops$ costs $F(\loops)=\Fbase(1+(\loops-1)\rho)$, with
$\rho=n_\midblock\Flayer/\Fbase$ and $\Flayer=3F_{\mathrm{block}}$
(Equation~\ref{eqn:step-cost}). Averaged over the loop-count distribution, the
cost per step depends only on the mean loop count,
\begin{equation}
\label{eqn:isoflop}
\frac{\Floop}{\Fbase} = 1 + (\Es-1)\,\rho .
\end{equation}

For the \texttt{small} model with $n=12$, $d=768$, $L=1024$, $n_\midblock=2$,
and $\loopmax=12$, we get $\rho=0.0956$ under Chinchilla and $\rho=0.167$
under Kaplan. With the global batch fixed, a baseline budget of $M$ steps
equals $M\Fbase/\Floop$ looped steps (Table~\ref{tab:flopprop}).

\begin{figure}[t]
  \centering
  \includegraphics[width=\textwidth]
    {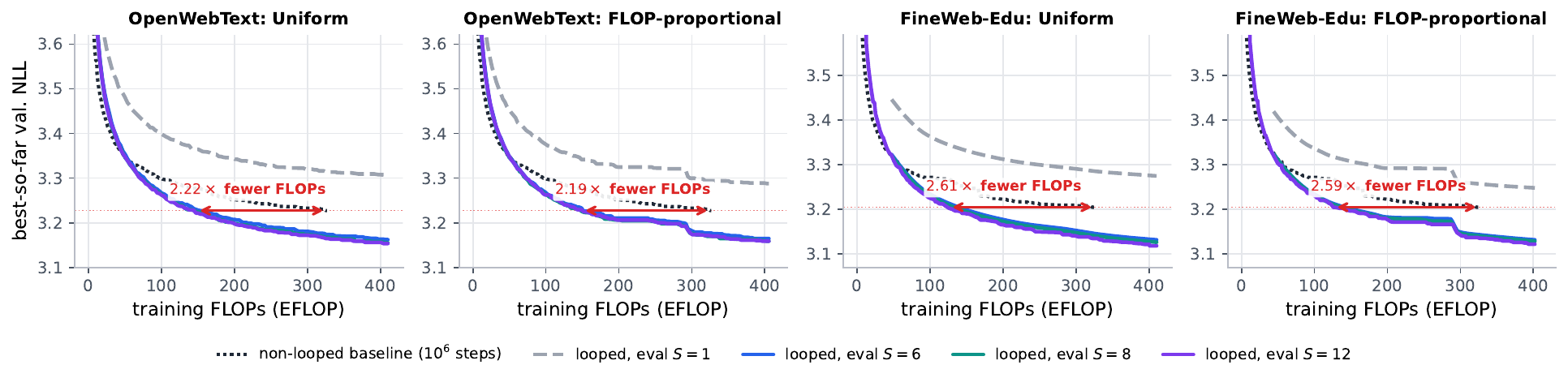}
  \caption{\textbf{Training curves under the Kaplan FLOP convention} (Chinchilla version in Figure~\ref{fig:training-elbow}).}
  \label{fig:kaplan-elbows}
\end{figure}

\paragraph{Chinchilla and Kaplan conventions.}
Scaling-law papers count compute in two ways \citep{pearce2024reconciling}.
Chinchilla \citep{hoffmann2022training} counts all parameters. Kaplan
\citep{kaplan2020scaling} does not count the input embedding and the output
layer, so it removes the two vocab terms in Table~\ref{tab:flop-components}.
Our main results use Chinchilla, which \citet{pearce2024reconciling} recommend
for small models. Figure~\ref{fig:kaplan-elbows} shows the same curves under
Kaplan. Without the vocab terms the baseline step is cheaper but a looped
block costs the same. This makes looping relatively costlier
(Table~\ref{tab:flopprop}).
\label{app:kaplan-curves}

\section{Training configuration and optimization}
\label{app:training-config}
\begin{figure}[t]
  \centering
  \includegraphics[width=\textwidth]
    {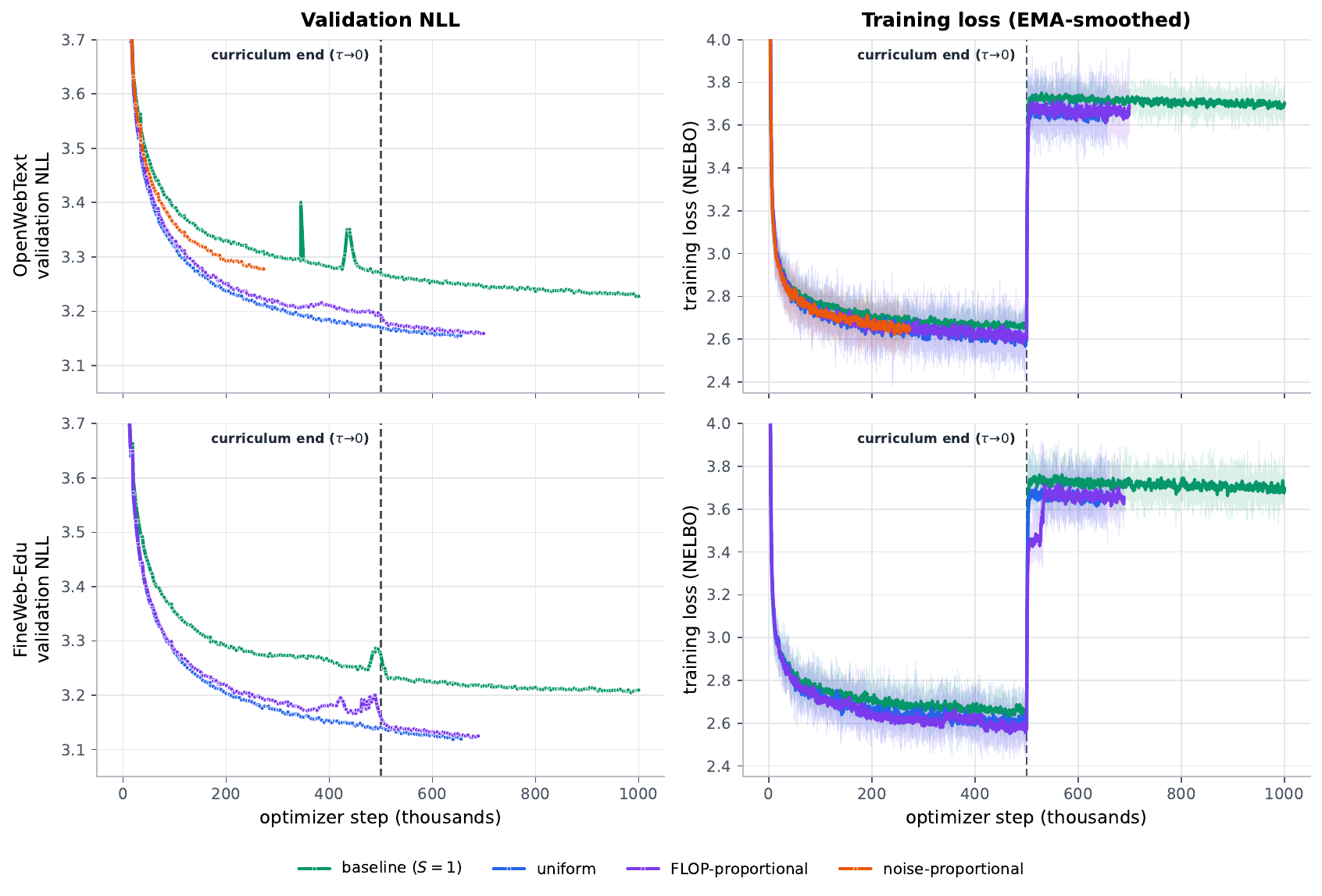}
  \caption{\textbf{Validation NLL and training loss by loop-count distribution on OWT
  (top) and FineWeb-Edu (bottom).} There is no noise-proportional run on FineWeb-Edu.
  \textbf{Left:} exact validation NLL against optimizer step.
  \textbf{Right:} EMA-smoothed training loss (NELBO). The dashed line marks
  the end of the curriculum at step \(5\times10^{5}\).}
  \label{fig:training-curves}
\end{figure}

Every run uses the same configuration
(Table~\ref{tab:trainconfig}). Figure~\ref{fig:training-curves} shows the
training curves.

\paragraph{Pretraining datasets.}
We train on OpenWebText (OWT)~\citep{Gokaslan2019OpenWeb} and on the
\texttt{sample-10BT} configuration of FineWeb-Edu~\citep{penedo2024fineweb}
(\(9{,}672{,}101\) documents, nominally \(10\)B tokens). In both cases we
hold out the last \(100{,}000\) documents for validation and train on the
rest. We tokenize with the GPT-2 tokenizer (\(V=50{,}257\)), put an EOS
between documents, and pack the stream into \(1{,}024\)-token sequences made
of BOS, \(1{,}022\) payload tokens, and EOS. Packing is done separately for
each batch of \(1{,}000\) documents, and the leftover tokens of each batch
are discarded. An epoch has \(8.94\)B training tokens on OWT and \(9.86\)B
on FineWeb-Edu, which is \(17{,}053\) and \(18{,}815\) optimizer steps with
\(512\) sequences per batch.

\paragraph{Curriculum and Gumbel sampling.}
We use the \methodpp{} efficient curriculum: the diffusion transformation
operator is approximated by a degree-9 polynomial (\texttt{poly9}), fit from its
\(150\)-term series over the curriculum window, and the tempered softmax input
is simulated from its top-\(k=2\) entries. For the first \(5\times10^{5}\)
optimizer steps, training uses this relaxed curriculum loss with Gumbel-softmax
temperature \(\tau=10^{-3}\) and log-SNR window \(\gamma\in[-3.55,-1.85]\)
(the code's parameterization of the curriculum window). After that, training
switches to the exact discrete NELBO, which causes the training-loss regime
change in Figure~\ref{fig:training-curves}.
Diffusion time is drawn with antithetic (stratified) sampling and
\(\varepsilon=10^{-3}\). Importance sampling and change-of-variables
are off. The noise-proportional distribution builds its per-shard noise
bands from this same stratification (\S\ref{sec:method-noiseprop}).

\begin{table}[t]
\centering
\footnotesize
\setlength{\tabcolsep}{4pt}
\begin{minipage}[t]{0.48\textwidth}\centering
\begin{tabular}{@{}ll@{}}
\toprule
\multicolumn{2}{@{}l}{\textit{Model}} \\
\midrule
Architecture & DiT \\
Hidden size & 768 \\
Attention heads & 12 \\
Blocks & 12 \\
Time-embedding size & 128 \\
MLP ratio & 4 \\
Dropout & 0.1 \\
Embeddings (input, output) & untied \\
Sequence length & 1,024 \\
Vocabulary & GPT-2 (50,257) \\
\midrule
\multicolumn{2}{@{}l}{\textit{Compute}} \\
\midrule
GPUs & H100 \\
Batch size & 512 \\
Precision & bf16 \\
Validation interval & 5,000 steps \\
\bottomrule
\end{tabular}
\end{minipage}\hfill
\begin{minipage}[t]{0.48\textwidth}\centering
\begin{tabular}{@{}ll@{}}
\toprule
\multicolumn{2}{@{}l}{\textit{Diffusion}} \\
\midrule
Objective & \method{} ELBO \\
Time & continuous, antithetic \\
Parameterization & mean \\
Curriculum & first 500k steps \\
Temperature $\tau$ & $10^{-3}$ \\
Log-SNR range & $[-3.55,-1.85]$ \\
Top-$k$ & 2 \\
$\Tmap$ approximation & degree-9 polynomial \\
\midrule
\multicolumn{2}{@{}l}{\textit{Optimization}} \\
\midrule
Optimizer & AdamW, $\beta=(0.9, 0.999)$ \\
Learning rate & $3\times10^{-4}$ \\
Warmup & 2,500 steps, then constant \\
Weight decay & 0 \\
Gradient clipping & 1.0 \\
EMA decay & 0.9999 \\
\bottomrule
\end{tabular}
\end{minipage}
\caption{\textbf{Training configuration for pretraining on OWT and FineWeb-Edu.}}
\label{tab:trainconfig}
\end{table}

\section{Loop architecture selection ablations}
\label{app:loop-ablation}
We train with the setup of Appendix~\ref{app:training-config}, sampling
\(\loops\sim\mathcal U\{1,\dots,\loopmax\}\) and looping two contiguous blocks,
for \(50\)k steps with seed \(1\). First we move the slice start from block
\(0\) to block \(10\) with \(\loopmax{=}12\), then fix the slice at blocks
\(1\)--\(2\) and try \(\loopmax\in\{4,8,12,16\}\)
(Table~\ref{tab:loop-layer-ablation}). Looping early blocks works best, with
blocks \(1\)--\(2\) giving the lowest NLL, and NLL stops improving at
\(\loopmax{=}12\).

\begin{table}[t]
\centering
\scriptsize
\setlength{\tabcolsep}{2.6pt}
\renewcommand{\arraystretch}{0.88}
\begin{tabular}{@{}cccccccccc@{}}
\toprule
Looped blocks & \(n_H\) & \(n_T\) & \(\loopmax\) & \(\loops{=}1\) & \(\loops{=}6\) & \(\loops{=}8\) & \(\loops{=}12\) & \(\loops{=}16\) & \(\loops{=}24\) \\
\midrule
\multicolumn{10}{@{}l}{\emph{(a) Position sweep at \(\loopmax{=}12\)}} \\
0--1 & 0 & 10 & 12 & 3.539 & 3.456 & \textbf{3.449} & 3.454 & -- & 3.475 \\
\rowcolor{plpurplelight!35}1--2 & 1 & 9 & 12 & 3.548 & 3.425 & \textbf{3.416} & 3.420 & -- & 3.447 \\
2--3 & 2 & 8 & 12 & 3.567 & 3.433 & \textbf{3.423} & 3.426 & -- & 3.451 \\
3--4 & 3 & 7 & 12 & 3.548 & 3.428 & \textbf{3.418} & 3.421 & -- & 3.447 \\
4--5 & 4 & 6 & 12 & 3.548 & 3.433 & \textbf{3.423} & 3.426 & -- & 3.455 \\
5--6 & 5 & 5 & 12 & 3.554 & 3.430 & \textbf{3.420} & 3.424 & -- & 3.455 \\
6--7 & 6 & 4 & 12 & 3.551 & 3.440 & \textbf{3.431} & 3.436 & -- & 3.473 \\
7--8 & 7 & 3 & 12 & 3.565 & 3.455 & \textbf{3.447} & 3.454 & -- & 3.498 \\
8--9 & 8 & 2 & 12 & 3.560 & 3.465 & \textbf{3.459} & 3.467 & -- & 3.513 \\
9--10 & 9 & 1 & 12 & 3.595 & 3.491 & \textbf{3.485} & 3.495 & -- & 3.541 \\
10--11 & 10 & 0 & 12 & 3.578 & 3.499 & \textbf{3.495} & 3.504 & -- & 3.530 \\
\midrule
\multicolumn{10}{@{}l}{\emph{(b) Training-cap sweep at blocks 1--2}} \\
1--2 & 1 & 9 & 4 & 3.491 & \textbf{3.453} & 3.460 & 3.497 & -- & 3.677 \\
1--2 & 1 & 9 & 8 & 3.519 & 3.425 & \textbf{3.422} & 3.433 & -- & 3.492 \\
\rowcolor{plpurplelight!35}1--2 & 1 & 9 & 12 & 3.548 & 3.425 & \textbf{3.416} & 3.420 & -- & 3.447 \\
1--2 & 1 & 9 & 16 & 3.614 & 3.435 & 3.424 & \textbf{3.416} & 3.417 & 3.431 \\
\bottomrule
\end{tabular}
\caption{\textbf{Choice of looped blocks and training cap} (\(\downarrow\)). Validation NLL after \(50\)k steps on OWT with the uniform loop-count distribution. Each run loops two consecutive blocks, with \(n_H\) blocks before them and \(n_T\) after. Bold marks the best loop count in each row, and the shaded row is the setting used in the paper.}
\label{tab:loop-layer-ablation}
\end{table}

\section{Generation results}
\label{app:generation}

For each setting we draw 512 samples of 1,024 tokens with nucleus sampling
($p=0.9$), using both the ancestral and the $\Psi$ sampler. Generative
perplexity is computed with GPT-2 Large and measures sample quality, and
unigram entropy measures diversity.
Table~\ref{tab:generation-ppl} gives the generative perplexity and
Table~\ref{tab:generation-entropy} the entropy behind
Figure~\ref{fig:reference-ppl-entropy}.

\begin{table*}[t]
\centering
\scriptsize
\setlength{\tabcolsep}{2.6pt}
\renewcommand{\arraystretch}{0.88}
\begin{tabular}{r r rrrrr rrrrr}
\toprule
& Base & \multicolumn{5}{c}{Uniform} & \multicolumn{5}{c}{FLOP-proportional} \\
\cmidrule(lr){3-7}\cmidrule(lr){8-12}
NFE & -- & \(\loops{=}1\) & \(\loops{=}6\) & \(\loops{=}8\) & \(\loops{=}12\) & Adaptive & \(\loops{=}1\) & \(\loops{=}6\) & \(\loops{=}8\) & \(\loops{=}12\) & Adaptive \\
\midrule
\multicolumn{12}{l}{\emph{Ancestral}} \\
32 & 45.54 & 43.91 & 42.59 & 42.60 & 43.02 & \underline{\textbf{41.73}} & 44.82 & 42.65 & 42.92 & 43.60 & \textbf{42.08} \\
64 & 40.40 & 39.87 & 37.98 & 37.60 & 37.98 & \underline{\textbf{37.37}} & 40.42 & \textbf{38.16} & 38.75 & 39.21 & 38.39 \\
128 & 38.17 & 38.63 & 35.61 & 35.22 & \underline{\textbf{34.98}} & 35.58 & 39.06 & 36.55 & \textbf{36.35} & 37.50 & 36.66 \\
256 & 37.52 & 37.49 & 34.23 & \underline{\textbf{33.95}} & 34.42 & 34.07 & 36.50 & \textbf{34.37} & 34.67 & 34.95 & 34.56 \\
512 & 36.94 & 37.44 & 34.17 & 33.98 & 34.16 & \underline{\textbf{33.35}} & 36.91 & 34.59 & \textbf{33.89} & 34.58 & 34.39 \\
1,024 & 36.06 & 36.31 & 32.71 & \underline{\textbf{32.53}} & 33.56 & 32.74 & 36.67 & 33.97 & 33.98 & 34.06 & \textbf{33.59} \\
2,048 & 36.17 & 35.02 & 33.02 & 32.88 & 32.82 & \underline{\textbf{32.80}} & 36.13 & 33.67 & 34.04 & 33.75 & \textbf{33.23} \\
4,096 & 34.83 & 35.77 & 33.11 & 33.50 & 33.29 & \underline{\textbf{33.07}} & 36.62 & 34.03 & 33.33 & 33.98 & \textbf{33.28} \\
\midrule
\multicolumn{12}{l}{\emph{\(\Psi\)-sampler}} \\
32 & 44.83 & 43.57 & 41.97 & 42.00 & 42.86 & \underline{\textbf{41.33}} & 44.54 & 42.36 & \textbf{42.14} & 42.88 & 42.14 \\
64 & 39.19 & 39.07 & 36.67 & 36.74 & 37.14 & \underline{\textbf{36.65}} & 39.31 & \textbf{37.08} & 37.58 & 38.11 & 37.77 \\
128 & 36.08 & 35.98 & \underline{\textbf{33.84}} & 34.00 & 34.02 & 34.22 & 36.77 & 35.11 & \textbf{34.71} & 34.99 & 34.84 \\
256 & 34.09 & 33.68 & 31.23 & \underline{\textbf{31.21}} & 31.92 & 31.44 & 33.16 & \textbf{31.76} & 31.98 & 32.36 & 31.81 \\
512 & 30.66 & 31.23 & 29.69 & 29.41 & 29.34 & \underline{\textbf{29.30}} & 31.10 & 29.69 & 29.74 & 29.75 & \textbf{29.60} \\
1,024 & 27.58 & 26.73 & 25.90 & 25.66 & 25.83 & \underline{\textbf{25.54}} & 26.90 & \textbf{26.12} & 26.21 & 26.45 & 26.20 \\
2,048 & 23.47 & 22.08 & 21.74 & 21.70 & 21.87 & \underline{\textbf{21.43}} & 22.06 & 21.78 & 21.95 & \textbf{21.76} & 21.84 \\
4,096 & 18.30 & 17.79 & \underline{\textbf{17.48}} & 17.78 & 17.75 & 17.62 & \textbf{17.54} & 17.90 & 18.44 & 18.30 & 17.83 \\
\bottomrule
\end{tabular}
\caption{\textbf{Generation perplexity across loop counts} (\(\downarrow\)). Bold marks the best loop count within each looped checkpoint, and underline marks the best value in the row.}
\label{tab:generation-ppl}
\end{table*}

\begin{table*}[t]
\centering
\scriptsize
\setlength{\tabcolsep}{2.6pt}
\renewcommand{\arraystretch}{0.88}
\begin{tabular}{r r rrrrr rrrrr}
\toprule
& Base & \multicolumn{5}{c}{Uniform} & \multicolumn{5}{c}{FLOP-proportional} \\
\cmidrule(lr){3-7}\cmidrule(lr){8-12}
NFE & -- & \(\loops{=}1\) & \(\loops{=}6\) & \(\loops{=}8\) & \(\loops{=}12\) & Adaptive & \(\loops{=}1\) & \(\loops{=}6\) & \(\loops{=}8\) & \(\loops{=}12\) & Adaptive \\
\midrule
\multicolumn{12}{l}{\emph{Ancestral}} \\
32 & 5.42 & 5.40 & 5.42 & 5.42 & \underline{\textbf{5.42}} & 5.41 & 5.40 & 5.41 & \textbf{5.42} & 5.41 & 5.41 \\
64 & 5.40 & 5.40 & \underline{\textbf{5.41}} & 5.40 & 5.41 & 5.41 & 5.38 & 5.40 & 5.41 & \textbf{5.41} & 5.41 \\
128 & 5.39 & 5.39 & 5.39 & 5.40 & 5.40 & \textbf{5.40} & 5.39 & 5.40 & 5.40 & \underline{\textbf{5.41}} & 5.40 \\
256 & 5.38 & 5.38 & 5.37 & 5.38 & \underline{\textbf{5.38}} & 5.38 & 5.36 & 5.37 & \textbf{5.38} & 5.38 & 5.38 \\
512 & 5.37 & 5.37 & \underline{\textbf{5.38}} & 5.38 & 5.37 & 5.37 & 5.35 & \textbf{5.37} & 5.36 & 5.36 & 5.36 \\
1,024 & 5.37 & 5.36 & 5.36 & 5.36 & \underline{\textbf{5.37}} & 5.37 & 5.36 & 5.37 & \textbf{5.37} & 5.37 & 5.36 \\
2,048 & \underline{5.37} & 5.34 & 5.35 & \textbf{5.36} & 5.35 & 5.35 & 5.33 & 5.36 & \textbf{5.36} & 5.36 & 5.35 \\
4,096 & 5.35 & 5.35 & 5.38 & \textbf{5.38} & 5.37 & 5.38 & 5.36 & \underline{\textbf{5.38}} & 5.37 & 5.37 & 5.37 \\
\midrule
\multicolumn{12}{l}{\emph{\(\Psi\)-sampler}} \\
32 & 5.41 & 5.40 & 5.42 & 5.41 & \underline{\textbf{5.42}} & 5.41 & 5.40 & \textbf{5.41} & 5.41 & 5.41 & 5.41 \\
64 & 5.40 & 5.39 & 5.40 & 5.39 & \textbf{5.40} & 5.40 & 5.38 & 5.40 & 5.40 & 5.40 & \underline{\textbf{5.40}} \\
128 & 5.38 & 5.38 & 5.39 & 5.39 & 5.39 & \underline{\textbf{5.39}} & 5.37 & 5.39 & \textbf{5.39} & 5.38 & 5.39 \\
256 & \underline{5.36} & 5.35 & 5.35 & 5.36 & \textbf{5.36} & 5.36 & 5.34 & 5.36 & 5.36 & 5.36 & \textbf{5.36} \\
512 & 5.34 & 5.33 & \underline{\textbf{5.35}} & 5.34 & 5.34 & 5.34 & 5.32 & 5.33 & 5.34 & 5.33 & \textbf{5.34} \\
1,024 & 5.31 & 5.29 & 5.32 & \underline{\textbf{5.32}} & 5.32 & 5.31 & 5.28 & \textbf{5.31} & 5.31 & 5.31 & 5.31 \\
2,048 & \underline{5.27} & 5.22 & \textbf{5.25} & 5.25 & 5.25 & 5.24 & 5.21 & 5.25 & 5.24 & 5.24 & \textbf{5.25} \\
4,096 & 5.18 & 5.14 & 5.19 & 5.19 & 5.18 & \textbf{5.19} & 5.14 & 5.20 & \underline{\textbf{5.22}} & 5.21 & 5.20 \\
\bottomrule
\end{tabular}
\caption{\textbf{Unigram entropy of generated samples across loop counts} (\(\uparrow\)). Bold marks the best loop count within each looped checkpoint, and underline marks the best value in the row.}
\label{tab:generation-entropy}
\end{table*}

\FloatBarrier
\section{Zero-shot likelihood results}
\label{app:zero-shot}
\label{app:zero-shot-ppl}

We evaluate on the datasets in Table~\ref{tab:zero-shot-data}. We use the EMA weights of each checkpoint. Documents are packed into
1,024-token sequences \([\mathrm{BOS},x_{1:1022},\mathrm{EOS}]\), and all
1,024 positions are scored.

\begin{table*}[t]
\centering
\scriptsize
\setlength{\tabcolsep}{3.5pt}
\begin{minipage}[b]{0.47\textwidth}\centering
\begin{tabular}{llrr}
\toprule
Dataset & Split & Documents & Tokens \\
\midrule
PTB & validation & 3,370 & 75,776 \\
WikiText-103 & validation & 3,760 & 233,472 \\
LM1B & test & 306,688 & 8,475,648 \\
LAMBADA & test & 5,153 & 421,888 \\
AG News & test & 7,600 & 389,120 \\
PubMed & validation & 6,633 & 26,807,296 \\
arXiv & validation & 6,436 & 53,453,824 \\
\bottomrule
\end{tabular}
\end{minipage}\hfill
\begin{minipage}[b]{0.47\textwidth}\centering
\begin{tabular}{lrrrr}
\toprule
& \multicolumn{2}{c}{Uniform} & \multicolumn{2}{c}{FLOP-proportional} \\
\cmidrule(lr){2-3}\cmidrule(lr){4-5}
Dataset & Loops & Saved & Loops & Saved \\
\midrule
PTB & 6.18 & 48.5\% & 6.14 & 48.8\% \\
WikiText-103 & 5.84 & 51.3\% & 5.67 & 52.8\% \\
LM1B & 6.13 & 48.9\% & 5.89 & 50.9\% \\
LAMBADA & 5.48 & 54.3\% & 5.43 & 54.8\% \\
AG News & 6.63 & 44.8\% & 6.37 & 46.9\% \\
PubMed & 6.35 & 47.1\% & 6.06 & 49.5\% \\
arXiv & 6.35 & 47.1\% & 6.15 & 48.8\% \\
\midrule
\textit{Average} & 6.14 & 48.9\% & 5.96 & 50.3\% \\
\bottomrule
\end{tabular}
\end{minipage}
\\
\begin{minipage}[t]{0.47\textwidth}\centering
\caption{\textbf{Likelihood datasets.}}
\label{tab:zero-shot-data}
\end{minipage}\hfill
\begin{minipage}[t]{0.47\textwidth}\centering
\caption{\textbf{Adaptive looping on likelihood datasets} ($\epsilon=0.1$, $S_{\max}=12$). Saved counts the loop iterations skipped compared to fixed $S=12$.}
\label{tab:zero-shot-ppl-adaptive}
\end{minipage}
\end{table*}

\begin{table*}[t]
\centering
\scriptsize
\setlength{\tabcolsep}{2.4pt}
\resizebox{\textwidth}{!}{%
\begin{tabular}{lrrrrrrrrrrr}
\toprule
& & \multicolumn{5}{c}{Uniform} & \multicolumn{5}{c}{FLOP-proportional} \\
\cmidrule(lr){3-7}\cmidrule(lr){8-12}
& Base & $S{=}1$ & $S{=}6$ & $S{=}8$ & $S{=}12$ & Adaptive & $S{=}1$ & $S{=}6$ & $S{=}8$ & $S{=}12$ & Adaptive \\
\midrule
PTB & 94.86{\tiny$\,\pm$10.87} & 105.63{\tiny$\,\pm$11.81} & 85.18{\tiny$\,\pm$9.07} & 84.01{\tiny$\,\pm$8.88} & \underline{\textbf{83.97{\tiny$\,\pm$8.60}}} & 85.14{\tiny$\,\pm$9.01} & 103.87{\tiny$\,\pm$11.39} & 85.47{\tiny$\,\pm$9.65} & \textbf{85.19{\tiny$\,\pm$9.77}} & 85.55{\tiny$\,\pm$9.56} & 85.43{\tiny$\,\pm$9.68} \\
WikiText-103 & 34.46{\tiny$\,\pm$1.11} & 37.74{\tiny$\,\pm$1.26} & 31.70{\tiny$\,\pm$0.93} & 31.39{\tiny$\,\pm$0.94} & \underline{\textbf{31.31{\tiny$\,\pm$0.92}}} & 31.75{\tiny$\,\pm$0.89} & 36.96{\tiny$\,\pm$1.30} & 31.65{\tiny$\,\pm$0.95} & 31.46{\tiny$\,\pm$0.91} & \textbf{31.44{\tiny$\,\pm$0.91}} & 31.71{\tiny$\,\pm$0.95} \\
LM1B & 74.21{\tiny$\,\pm$0.27} & 81.30{\tiny$\,\pm$0.31} & 71.16{\tiny$\,\pm$0.26} & 70.59{\tiny$\,\pm$0.27} & \textbf{70.38{\tiny$\,\pm$0.26}} & 71.22{\tiny$\,\pm$0.27} & 78.64{\tiny$\,\pm$0.37} & 70.35{\tiny$\,\pm$0.33} & \underline{\textbf{70.05{\tiny$\,\pm$0.32}}} & 70.15{\tiny$\,\pm$0.32} & 70.47{\tiny$\,\pm$0.33} \\
LAMBADA & 51.83{\tiny$\,\pm$0.65} & 53.68{\tiny$\,\pm$0.88} & 48.02{\tiny$\,\pm$0.72} & 47.76{\tiny$\,\pm$0.78} & \textbf{47.70{\tiny$\,\pm$0.86}} & 48.21{\tiny$\,\pm$0.71} & 52.65{\tiny$\,\pm$0.77} & 46.38{\tiny$\,\pm$1.04} & \underline{\textbf{46.33{\tiny$\,\pm$1.07}}} & 46.45{\tiny$\,\pm$1.12} & 46.56{\tiny$\,\pm$1.04} \\
AG News & 69.95{\tiny$\,\pm$1.73} & 80.98{\tiny$\,\pm$2.19} & 59.21{\tiny$\,\pm$1.53} & 57.93{\tiny$\,\pm$1.49} & \underline{\textbf{57.31{\tiny$\,\pm$1.54}}} & 58.68{\tiny$\,\pm$1.52} & 80.41{\tiny$\,\pm$2.43} & 59.88{\tiny$\,\pm$1.50} & 58.77{\tiny$\,\pm$1.42} & \textbf{58.35{\tiny$\,\pm$1.49}} & 59.63{\tiny$\,\pm$1.50} \\
PubMed & 44.91{\tiny$\,\pm$0.10} & 48.28{\tiny$\,\pm$0.09} & 38.91{\tiny$\,\pm$0.06} & 38.30{\tiny$\,\pm$0.06} & \underline{\textbf{37.95{\tiny$\,\pm$0.06}}} & 38.82{\tiny$\,\pm$0.06} & 47.21{\tiny$\,\pm$0.09} & 39.16{\tiny$\,\pm$0.06} & 38.52{\tiny$\,\pm$0.05} & \textbf{38.11{\tiny$\,\pm$0.05}} & 39.23{\tiny$\,\pm$0.06} \\
arXiv & 39.78{\tiny$\,\pm$0.12} & 42.98{\tiny$\,\pm$0.13} & 35.02{\tiny$\,\pm$0.10} & 34.47{\tiny$\,\pm$0.10} & \underline{\textbf{34.13{\tiny$\,\pm$0.10}}} & 34.97{\tiny$\,\pm$0.10} & 42.57{\tiny$\,\pm$0.13} & 35.27{\tiny$\,\pm$0.10} & 34.71{\tiny$\,\pm$0.10} & \textbf{34.37{\tiny$\,\pm$0.10}} & 35.31{\tiny$\,\pm$0.10} \\
\bottomrule
\end{tabular}}
\caption{\textbf{Zero-shot likelihood perplexity across loop counts} ($\downarrow$). Mean and standard deviation over five seeds. Adaptive looping uses $S_{\max}=12$. Bold marks the best loop count within each looped checkpoint, and underline marks the best value in the row.}
\label{tab:zero-shot-ppl-all}
\end{table*}

Table~\ref{tab:zero-shot-ppl-all} reports every configuration, and
Table~\ref{tab:zero-shot-ppl-adaptive} gives the average adaptive loop count
and the share of loop iterations saved on each dataset.

\label{app:single-call}
Figures~\ref{fig:adaptive-stopping} and~\ref{fig:loop-reliability} use up to
1,024 blocks from OWT validation and from each dataset above. Each block is
corrupted once at \(t\in\{0.1,0.3,0.5,0.7,0.9\}\) and scored after every loop
count \(\loops=1,\ldots,12\). Adaptive looping is replayed offline with
\(\epsilon=0.1\). The curves average the per-token NLL of each dataset with
equal weight and report the exponential of that average. In
Figure~\ref{fig:loop-reliability}c, bins with fewer than \(300\) tokens in
more than two datasets are omitted.

\section{Multiple-choice results}
\label{app:zero-shot-mcq}

We use ten commonsense and reasoning multiple-choice tasks
(Table~\ref{tab:mcq-tasks}) with the task definitions from
\texttt{lm-eval-harness} 0.4.11~\citep{eval-harness}, default prompts, and
no few-shot examples. Tables~\ref{tab:zero-shot-mcq-results}
and~\ref{tab:zero-shot-mcq-adaptive} give the full results.

\begin{table}[t]
\centering
\scriptsize
\setlength{\tabcolsep}{3.5pt}
\begin{tabular}{llrr}
\toprule
Task & Split & Questions & Choices \\
\midrule
ARC-Easy & test & 2,376 & 9,501 \\
ARC-Challenge & test & 1,172 & 4,687 \\
HellaSwag & validation & 10,042 & 40,168 \\
WinoGrande & validation & 1,267 & 2,534 \\
PIQA & validation & 1,838 & 3,676 \\
MathQA & test & 2,985 & 14,925 \\
OpenBookQA & test & 500 & 2,000 \\
BoolQ & validation & 3,270 & 6,540 \\
RACE-High & test & 1,045 & 4,180 \\
Social IQa & validation & 1,954 & 5,862 \\
\bottomrule
\end{tabular}
\caption{\textbf{Multiple-choice tasks.}}
\label{tab:mcq-tasks}
\end{table}

\begin{table*}[t]
\centering
\scriptsize
\setlength{\tabcolsep}{2.4pt}
\resizebox{\textwidth}{!}{%
\begin{tabular}{lrrrrrrrrrrr}
\toprule
& & \multicolumn{5}{c}{Uniform} & \multicolumn{5}{c}{FLOP-proportional} \\
\cmidrule(lr){3-7}\cmidrule(lr){8-12}
& Base & $S{=}1$ & $S{=}6$ & $S{=}8$ & $S{=}12$ & Adaptive & $S{=}1$ & $S{=}6$ & $S{=}8$ & $S{=}12$ & Adaptive \\
\midrule
ARC-Easy & 28.66{\tiny$\,\pm$0.37} & \textbf{29.08{\tiny$\,\pm$0.47}} & 27.89{\tiny$\,\pm$0.69} & 27.73{\tiny$\,\pm$0.34} & 27.55{\tiny$\,\pm$0.31} & 27.83{\tiny$\,\pm$0.16} & \underline{\textbf{29.17{\tiny$\,\pm$0.37}}} & 28.16{\tiny$\,\pm$0.61} & 28.43{\tiny$\,\pm$0.75} & 28.62{\tiny$\,\pm$0.85} & 28.70{\tiny$\,\pm$0.62} \\
ARC-Challenge & 25.26{\tiny$\,\pm$0.75} & 25.97{\tiny$\,\pm$0.99} & 26.60{\tiny$\,\pm$1.10} & 26.42{\tiny$\,\pm$0.60} & \textbf{26.84{\tiny$\,\pm$0.96}} & 26.52{\tiny$\,\pm$0.73} & 25.97{\tiny$\,\pm$0.74} & \underline{\textbf{27.30{\tiny$\,\pm$1.37}}} & 26.59{\tiny$\,\pm$0.67} & 26.18{\tiny$\,\pm$0.24} & 26.23{\tiny$\,\pm$0.84} \\
HellaSwag & 25.49{\tiny$\,\pm$0.09} & 25.73{\tiny$\,\pm$0.21} & 27.94{\tiny$\,\pm$0.25} & 28.31{\tiny$\,\pm$0.23} & \textbf{28.62{\tiny$\,\pm$0.21}} & 28.42{\tiny$\,\pm$0.16} & 26.56{\tiny$\,\pm$0.17} & 28.51{\tiny$\,\pm$0.20} & 28.76{\tiny$\,\pm$0.15} & \underline{\textbf{29.11{\tiny$\,\pm$0.12}}} & 28.91{\tiny$\,\pm$0.16} \\
WinoGrande & 50.26{\tiny$\,\pm$1.33} & 49.80{\tiny$\,\pm$1.49} & \textbf{50.32{\tiny$\,\pm$1.75}} & 50.09{\tiny$\,\pm$1.62} & 49.85{\tiny$\,\pm$2.08} & 50.24{\tiny$\,\pm$1.81} & 50.58{\tiny$\,\pm$2.05} & \underline{\textbf{50.83{\tiny$\,\pm$1.49}}} & 50.69{\tiny$\,\pm$0.91} & 50.23{\tiny$\,\pm$1.33} & 49.98{\tiny$\,\pm$1.22} \\
PIQA & 51.37{\tiny$\,\pm$1.18} & 51.53{\tiny$\,\pm$0.84} & 51.39{\tiny$\,\pm$0.44} & 51.74{\tiny$\,\pm$0.52} & 52.01{\tiny$\,\pm$1.03} & \textbf{52.07{\tiny$\,\pm$0.93}} & 52.73{\tiny$\,\pm$1.35} & 52.97{\tiny$\,\pm$1.09} & \underline{\textbf{53.26{\tiny$\,\pm$0.96}}} & 53.01{\tiny$\,\pm$0.91} & 53.06{\tiny$\,\pm$1.00} \\
MathQA & 20.89{\tiny$\,\pm$0.55} & 20.32{\tiny$\,\pm$1.17} & 20.72{\tiny$\,\pm$0.75} & 20.87{\tiny$\,\pm$0.93} & \underline{\textbf{20.98{\tiny$\,\pm$1.08}}} & 20.90{\tiny$\,\pm$0.74} & 20.60{\tiny$\,\pm$0.23} & \textbf{20.84{\tiny$\,\pm$0.32}} & 20.71{\tiny$\,\pm$0.13} & 20.64{\tiny$\,\pm$0.37} & 20.73{\tiny$\,\pm$0.43} \\
OpenBookQA & 28.04{\tiny$\,\pm$1.68} & 27.40{\tiny$\,\pm$2.07} & \textbf{28.36{\tiny$\,\pm$1.66}} & 27.48{\tiny$\,\pm$1.70} & 27.84{\tiny$\,\pm$1.81} & 27.48{\tiny$\,\pm$1.37} & 27.72{\tiny$\,\pm$1.19} & 30.48{\tiny$\,\pm$0.92} & 30.84{\tiny$\,\pm$0.85} & 30.68{\tiny$\,\pm$1.07} & \underline{\textbf{31.04{\tiny$\,\pm$1.19}}} \\
BoolQ & 49.31{\tiny$\,\pm$0.91} & 50.13{\tiny$\,\pm$0.66} & \textbf{51.06{\tiny$\,\pm$0.75}} & 51.03{\tiny$\,\pm$0.86} & 50.65{\tiny$\,\pm$1.27} & 50.95{\tiny$\,\pm$0.91} & \underline{\textbf{53.31{\tiny$\,\pm$0.68}}} & 50.71{\tiny$\,\pm$1.19} & 50.91{\tiny$\,\pm$0.93} & 50.65{\tiny$\,\pm$0.91} & 50.84{\tiny$\,\pm$0.83} \\
RACE-High & 27.81{\tiny$\,\pm$2.14} & 27.43{\tiny$\,\pm$1.56} & 27.98{\tiny$\,\pm$1.98} & \textbf{28.17{\tiny$\,\pm$2.35}} & 27.98{\tiny$\,\pm$2.25} & 27.98{\tiny$\,\pm$2.08} & 27.25{\tiny$\,\pm$1.73} & 28.46{\tiny$\,\pm$2.27} & 28.46{\tiny$\,\pm$1.89} & \underline{\textbf{28.78{\tiny$\,\pm$1.64}}} & 28.33{\tiny$\,\pm$2.19} \\
Social IQa & 35.74{\tiny$\,\pm$1.15} & \underline{\textbf{36.25{\tiny$\,\pm$0.57}}} & 35.47{\tiny$\,\pm$0.37} & 35.70{\tiny$\,\pm$0.45} & 35.58{\tiny$\,\pm$0.51} & 35.90{\tiny$\,\pm$0.90} & 35.49{\tiny$\,\pm$0.79} & 36.03{\tiny$\,\pm$0.20} & \textbf{36.19{\tiny$\,\pm$0.40}} & 35.95{\tiny$\,\pm$0.76} & 36.03{\tiny$\,\pm$0.68} \\
\midrule
\textit{Average} & 34.28{\tiny$\,\pm$0.44} & 34.37{\tiny$\,\pm$0.49} & 34.77{\tiny$\,\pm$0.37} & 34.75{\tiny$\,\pm$0.38} & 34.79{\tiny$\,\pm$0.53} & \textbf{34.83{\tiny$\,\pm$0.40}} & 34.94{\tiny$\,\pm$0.44} & 35.43{\tiny$\,\pm$0.40} & \underline{\textbf{35.48{\tiny$\,\pm$0.41}}} & 35.39{\tiny$\,\pm$0.50} & 35.38{\tiny$\,\pm$0.56} \\
\bottomrule
\end{tabular}}
\caption{\textbf{Zero-shot multiple-choice accuracy across loop counts.} Mean and standard deviation over five seeds. Bold marks the best loop count within each looped checkpoint, and underline marks the best value in the row.}
\label{tab:zero-shot-mcq-results}
\end{table*}

\begin{table}[t]
\centering
\scriptsize
\setlength{\tabcolsep}{3.5pt}
\begin{tabular}{lrrrrrrrr}
\toprule
& \multicolumn{4}{c}{Uniform} & \multicolumn{4}{c}{FLOP-proportional} \\
\cmidrule(lr){2-5}\cmidrule(lr){6-9}
Task & Accuracy & Loops & Converged & Saved & Accuracy & Loops & Converged & Saved \\
\midrule
ARC-Easy & 27.83 & 9.61 & 77.4\% & 20.0\% & \textbf{28.70} & \textbf{8.71} & \textbf{88.3\%} & \textbf{27.4\%} \\
ARC-Challenge & \textbf{26.52} & 9.54 & 76.7\% & 20.5\% & 26.23 & \textbf{8.53} & \textbf{89.2\%} & \textbf{28.9\%} \\
HellaSwag & 28.42 & 8.49 & 89.3\% & 29.2\% & \textbf{28.91} & \textbf{7.30} & \textbf{97.0\%} & \textbf{39.2\%} \\
WinoGrande & \textbf{50.24} & 9.95 & 77.0\% & 17.1\% & 49.98 & \textbf{9.10} & \textbf{85.7\%} & \textbf{24.1\%} \\
PIQA & 52.07 & 9.43 & 79.8\% & 21.4\% & \textbf{53.06} & \textbf{8.50} & \textbf{89.8\%} & \textbf{29.2\%} \\
MathQA & \textbf{20.90} & 9.61 & 77.4\% & 19.9\% & 20.73 & \textbf{8.39} & \textbf{91.2\%} & \textbf{30.1\%} \\
OpenBookQA & 27.48 & 9.29 & 88.1\% & 22.6\% & \textbf{31.04} & \textbf{8.94} & \textbf{92.0\%} & \textbf{25.5\%} \\
BoolQ & \textbf{50.95} & 7.76 & 95.3\% & 35.3\% & 50.84 & \textbf{6.81} & \textbf{99.5\%} & \textbf{43.3\%} \\
RACE-High & 27.98 & 6.40 & 100.0\% & 46.7\% & \textbf{28.33} & \textbf{5.96} & \textbf{100.0\%} & \textbf{50.3\%} \\
Social IQa & 35.90 & 10.13 & 67.0\% & 15.6\% & \textbf{36.03} & \textbf{9.03} & \textbf{85.2\%} & \textbf{24.8\%} \\
\midrule
\textit{Average} & 34.83 & 8.89 & 84.3\% & 26.0\% & \textbf{35.38} & \textbf{7.82} & \textbf{93.7\%} & \textbf{34.8\%} \\
\bottomrule
\end{tabular}
\caption{\textbf{Adaptive looping on multiple-choice tasks} ($\epsilon=0.1$, $S_{\max}=12$). Saved counts the loop iterations skipped compared to fixed $S=12$, bold highlights the better checkpoint.}
\label{tab:zero-shot-mcq-adaptive}
\end{table}

Each answer choice is written as
\([\mathrm{BOS},\ \text{context},\ \text{answer},\ \mathrm{EOS}]\) and scored by
the token NLL averaged over all positions and \(1{,}024\) noise samples with
raw weights, following \methodpp{}~\citep{deschenaux2026diffusion}. The choice
with the lowest NLL is picked, without length normalization. Choices longer
than \(1{,}024\) tokens get log-likelihood \(0\) (two BoolQ and 24 RACE-High
choices).

\begin{table*}[t]
\centering
\scriptsize
\setlength{\tabcolsep}{2.4pt}
\resizebox{\textwidth}{!}{%
\begin{tabular}{lrrrrrrrrrrr}
\toprule
& & \multicolumn{5}{c}{Uniform} & \multicolumn{5}{c}{FLOP-proportional} \\
\cmidrule(lr){3-7}\cmidrule(lr){8-12}
& Base & $S{=}1$ & $S{=}6$ & $S{=}8$ & $S{=}12$ & Adaptive & $S{=}1$ & $S{=}6$ & $S{=}8$ & $S{=}12$ & Adaptive \\
\midrule
ARC-Easy & 30.33{\tiny$\,\pm$0.26} & \textbf{31.58{\tiny$\,\pm$1.09}} & 30.72{\tiny$\,\pm$0.82} & 30.56{\tiny$\,\pm$0.61} & 30.53{\tiny$\,\pm$0.71} & 30.42{\tiny$\,\pm$0.58} & \underline{\textbf{33.35{\tiny$\,\pm$0.81}}} & 31.41{\tiny$\,\pm$0.64} & 31.58{\tiny$\,\pm$0.72} & 31.55{\tiny$\,\pm$0.81} & 31.64{\tiny$\,\pm$0.53} \\
ARC-Challenge & 26.43{\tiny$\,\pm$1.29} & 26.18{\tiny$\,\pm$0.64} & 26.83{\tiny$\,\pm$1.00} & 26.83{\tiny$\,\pm$1.52} & 27.05{\tiny$\,\pm$1.51} & \textbf{27.17{\tiny$\,\pm$1.51}} & 26.79{\tiny$\,\pm$1.46} & 27.34{\tiny$\,\pm$0.97} & \underline{\textbf{27.54{\tiny$\,\pm$0.67}}} & 27.46{\tiny$\,\pm$0.68} & 27.44{\tiny$\,\pm$0.54} \\
HellaSwag & 29.59{\tiny$\,\pm$0.13} & 28.92{\tiny$\,\pm$0.12} & 30.58{\tiny$\,\pm$0.18} & 30.72{\tiny$\,\pm$0.22} & \textbf{30.79{\tiny$\,\pm$0.19}} & 30.64{\tiny$\,\pm$0.21} & 30.26{\tiny$\,\pm$0.11} & 30.83{\tiny$\,\pm$0.12} & 30.84{\tiny$\,\pm$0.19} & \underline{\textbf{30.84{\tiny$\,\pm$0.20}}} & 30.71{\tiny$\,\pm$0.19} \\
WinoGrande & 50.01{\tiny$\,\pm$0.68} & 50.70{\tiny$\,\pm$1.02} & 50.31{\tiny$\,\pm$0.97} & 50.70{\tiny$\,\pm$1.10} & \textbf{50.73{\tiny$\,\pm$1.40}} & 50.61{\tiny$\,\pm$0.82} & 50.37{\tiny$\,\pm$1.46} & 50.70{\tiny$\,\pm$2.13} & 50.83{\tiny$\,\pm$1.74} & 50.75{\tiny$\,\pm$1.73} & \underline{\textbf{50.89{\tiny$\,\pm$2.19}}} \\
PIQA & 52.83{\tiny$\,\pm$0.59} & \textbf{53.82{\tiny$\,\pm$1.38}} & 52.64{\tiny$\,\pm$0.96} & 52.80{\tiny$\,\pm$1.14} & 52.76{\tiny$\,\pm$1.26} & 52.83{\tiny$\,\pm$1.46} & \underline{\textbf{54.89{\tiny$\,\pm$1.51}}} & 53.78{\tiny$\,\pm$1.43} & 53.93{\tiny$\,\pm$1.32} & 53.70{\tiny$\,\pm$1.52} & 53.93{\tiny$\,\pm$1.59} \\
MathQA & 20.19{\tiny$\,\pm$0.43} & 20.82{\tiny$\,\pm$0.56} & 21.24{\tiny$\,\pm$0.39} & 21.48{\tiny$\,\pm$0.50} & \underline{\textbf{21.51{\tiny$\,\pm$0.53}}} & 21.37{\tiny$\,\pm$0.41} & 21.18{\tiny$\,\pm$0.54} & 20.98{\tiny$\,\pm$0.30} & 20.84{\tiny$\,\pm$0.46} & \textbf{21.19{\tiny$\,\pm$0.29}} & 21.10{\tiny$\,\pm$0.55} \\
OpenBookQA & 31.16{\tiny$\,\pm$1.49} & 30.60{\tiny$\,\pm$0.86} & \textbf{32.00{\tiny$\,\pm$1.32}} & 31.36{\tiny$\,\pm$1.98} & 31.36{\tiny$\,\pm$0.96} & 31.88{\tiny$\,\pm$1.17} & \underline{\textbf{32.12{\tiny$\,\pm$0.76}}} & 31.24{\tiny$\,\pm$1.89} & 31.72{\tiny$\,\pm$1.81} & 31.48{\tiny$\,\pm$1.69} & 31.24{\tiny$\,\pm$1.94} \\
BoolQ & 45.66{\tiny$\,\pm$0.89} & 49.68{\tiny$\,\pm$0.67} & 50.04{\tiny$\,\pm$0.78} & \textbf{50.18{\tiny$\,\pm$0.49}} & 50.16{\tiny$\,\pm$0.65} & 49.88{\tiny$\,\pm$0.52} & 49.30{\tiny$\,\pm$1.17} & 51.46{\tiny$\,\pm$1.06} & 51.47{\tiny$\,\pm$0.79} & \underline{\textbf{51.72{\tiny$\,\pm$0.57}}} & 51.43{\tiny$\,\pm$0.80} \\
RACE-High & 28.29{\tiny$\,\pm$2.23} & 27.87{\tiny$\,\pm$1.52} & 28.80{\tiny$\,\pm$2.40} & 28.90{\tiny$\,\pm$2.26} & \textbf{29.17{\tiny$\,\pm$2.08}} & 28.78{\tiny$\,\pm$2.08} & 28.63{\tiny$\,\pm$2.18} & 29.07{\tiny$\,\pm$2.20} & 29.21{\tiny$\,\pm$2.24} & \underline{\textbf{29.30{\tiny$\,\pm$2.15}}} & 29.26{\tiny$\,\pm$2.26} \\
Social IQa & 35.46{\tiny$\,\pm$0.29} & 35.67{\tiny$\,\pm$0.31} & \textbf{35.90{\tiny$\,\pm$0.40}} & 35.78{\tiny$\,\pm$0.32} & 35.86{\tiny$\,\pm$0.52} & 35.79{\tiny$\,\pm$0.39} & \underline{\textbf{36.44{\tiny$\,\pm$0.34}}} & 36.23{\tiny$\,\pm$0.45} & 35.86{\tiny$\,\pm$0.55} & 35.72{\tiny$\,\pm$0.62} & 35.83{\tiny$\,\pm$0.37} \\
\midrule
\textit{Average} & 34.99{\tiny$\,\pm$0.31} & 35.58{\tiny$\,\pm$0.33} & 35.91{\tiny$\,\pm$0.21} & 35.93{\tiny$\,\pm$0.25} & \textbf{35.99{\tiny$\,\pm$0.37}} & 35.94{\tiny$\,\pm$0.23} & 36.33{\tiny$\,\pm$0.30} & 36.31{\tiny$\,\pm$0.28} & \underline{\textbf{36.38{\tiny$\,\pm$0.20}}} & 36.37{\tiny$\,\pm$0.22} & 36.35{\tiny$\,\pm$0.26} \\
\bottomrule
\end{tabular}}
\caption{\textbf{Zero-shot multiple-choice accuracy on FineWeb-Edu.} Mean and standard deviation over five seeds. Adaptive looping ($S_{\max}=12$) uses about 8 loop iterations per likelihood sample. Bold marks the best loop count within each looped checkpoint, and underline marks the best value in the row.}
\label{tab:fineweb-edu-mcq}
\end{table*}

We repeat the same experiment on the FineWeb-Edu checkpoints, with results in
Table~\ref{tab:fineweb-edu-mcq}. Adaptive looping uses on average \(8.02\)
loop iterations for the uniform checkpoint and \(7.99\) for the
FLOP-proportional checkpoint, which saves \(33\%\) and \(34\%\) of the loop
iterations relative to fixed \(S=12\).

\end{document}